\documentclass{article} % For LaTeX2e
\usepackage{iclr2026_conference,times}

\usepackage{amssymb}
\usepackage{mathtools}
\usepackage{amsmath}

\usepackage{minted}
\input{colorful.style.minted}
\usepackage{float}
\newfloat{listing}{h}{lop}
\floatname{listing}{Listing}

\usepackage{hyperref}
\usepackage{url}

\usepackage{booktabs}
\usepackage{graphicx}
\usepackage{subfigure}
\usepackage{caption}
\usepackage{tikz}
\usepackage{adjustbox}
\usepackage{enumitem}
\usepackage{comment}
\usepackage[capitalize,noabbrev]{cleveref}
\usepackage{textcomp}
\usepackage{xcolor}
\usepackage{subcaption} 
\usepackage{wrapfig}
\usepackage{makecell}
\usepackage[textsize=tiny]{todonotes}

\usepackage{algorithm}
\usepackage{multirow}

\usepackage[framemethod=TikZ]{mdframed}
\definecolor{lightGray}{gray}{0.97}
\definecolor{midGray}{gray}{0.75}
\definecolor{lightYellow}{RGB}{254,254,239}
\definecolor{lighterYellow}{RGB}{254,254,245}
\definecolor{darkGray}{rgb}{0.2,0.2,0.2}
\definecolor{darkerGray}{rgb}{0.3,0.3,0.3}
\mdfdefinestyle{myFrameStyle}{%
  linecolor=midGray,
  linewidth=0.5pt,
  roundcorner=3pt,
  skipabove=5pt,
  skipbelow=0pt,
  innertopmargin=5pt,
  innerbottommargin=5pt,
  innerrightmargin=5pt,
  innerleftmargin=5pt,
  leftmargin=0pt,
  rightmargin=0pt,
  backgroundcolor=lighterYellow
}
\mdfdefinestyle{frameStyleQuestion}{%
  linecolor=black!50,
  linewidth=0.5pt,
  roundcorner=3pt,
  skipabove=5pt,
  skipbelow=0pt,
  innertopmargin=2.5pt,
  innerbottommargin=2.5pt,
  innerrightmargin=3pt,
  innerleftmargin=3pt,
  leftmargin=0pt,
  rightmargin=0pt,
  backgroundcolor=lightYellow
}
\mdfdefinestyle{frameStyleListing}{%
  skipabove=0pt,skipbelow=0pt,
  linecolor=darkGray,
  linewidth=0.5pt,
  roundcorner=0pt,
  innertopmargin=5pt,
  innerbottommargin=5pt,
  backgroundcolor=lightGray
}

\newcommand{\frameQuestion}[1]{
    \vspace{-1pt}
    \begin{mdframed}[style=frameStyleQuestion,userdefinedwidth=
    \linewidth,align=center,skipabove=6pt,skipbelow=0pt]
    \centering{#1}
    \end{mdframed}
    \vspace{-2pt}
}

\newcommand{\myFrame}[1]{
    \vspace{4pt}
    \begin{mdframed}[style=myFrameStyle,userdefinedwidth=
    \linewidth,align=center,skipabove=6pt,skipbelow=0pt]
    {#1}
    \end{mdframed}\vspace{-8pt}
}

\renewcommand{\paragraph}[1]{\vspace{2pt}\noindent\textbf{#1}} % Bold paragraph titles

\usepackage[hang,flushmargin]{footmisc} % No indentation of footnotes, completely unnecessarily and eats up space

\newcommand{\liang}[1]{} % Hide comments

\definecolor{myBlue}{RGB}{10,10,230}

\newcommand{\algComment}[1]{{~~\textit{\color{black!35}#1}}}

\usepackage{siunitx} % For scientific notation
\usepackage[percent]{overpic}
\usepackage{pifont} % Symbols for plot legends
\usepackage{array} % For better vertical alignment
\usepackage{makecell}
\def\bx{{\boldsymbol{x}}}
\def\bs{{\boldsymbol{s}}}
\def\bK{{\mathbf{K}}} % Upright K
\def\bQ{{\mathbf{Q}}} % Upright Q
\def\bV{{\mathbf{V}}} % Upright V
\def\bS{{\mathbf{S}}} % Upright S
\def\bb{{\boldsymbol{b}}}
\def\bW{{\boldsymbol{W}}}
\def\btheta{{\boldsymbol{\theta}}}
\makeatletter
\newcommand{\shorteq}{\mathrel{\mkern0.2mu\mathpalette\shorteq@\relax\mkern0.2mu}}
\newcommand{\shorteq@}[2]{\scalebox{0.5}[1]{$\m@th#1=$}}

\makeatother
\usepackage{bbold} % For indicator function 1(...)
\usepackage{nicefrac}
\newcommand\lldots{\ifmmode$\lldots$\else\thinspace\makebox[1em][c]{.\hfil.\hfil.}\fi} % Shortly-spaced ldots

\definecolor{customblue}{HTML}{1A1AFF}
\definecolor{customred}{HTML}{FF4D4D}
\definecolor{darkYellow}{rgb}{0.8, 0.6, 0}

\makeatletter
\newcommand*{\rom}[1]{\expandafter\@slowromancap\romannumeral #1@}
\makeatother

\renewcommand{\sectionautorefname}{Section}
\renewcommand{\subsectionautorefname}{Section}
\renewcommand{\subsubsectionautorefname}{Section}

\newcommand{\refApp}[1]{\hyperref[#1]{Appendix\;\ref*{#1}}}    

\title{Can Transformers Really Do It All?\\On the Compatibility of Inductive\\Biases Across Tasks}
\author{%
    Damien Teney$^{1,3}$~~
    Liangze Jiang$^{1,2}$~~
    Hemanth Saratchandran$^3$~~
    Simon Lucey$^3$\\
    {$^1$}Idiap Research Institute~~
    {$^2$}EPFL~~
    {$^3$}Adelaide University\vspace{-2pt}%
}
\iclrfinalcopy % Uncomment for camera-ready version, but NOT for submission.
\begin{document}
\vspace{-3pt}
\maketitle
\vspace{-4pt}
\begin{abstract}
Transformers are remarkably versatile
and their design is largely consistent across a variety of applications.
But are they optimal for any given task or dataset?
The answer may be key for pushing AI beyond merely scaling current designs.

\vspace{2pt}
\textbf{Method.}
We present a method to optimize a transformer architecture for a given dataset,
which we use as a tool to study optimal task-specific inductive biases.
This method replaces the most important non-linearities (GeLUs,\;softmax) with functions learned on held-out data.
We then train the resulting architectures on other datasets,
as a way to evaluate the compatibility between pairs of tasks.

\vspace{2pt}
\textbf{Findings.}
On algorithmic toy tasks, we identify new architectures with dramatic improvements in learning speed,
in- and out-of-distribution generalization, and stability across seeds.
The new designs prove very task-specific however,
and indicate
that these tasks require inductive biases
very different from those of standard transformers.
On code and language modeling datasets,
we also find architectures with consistent, yet smaller improvements.
These designs transfer much better across datasets
and domains (English \& computer code).

\vspace{2pt}
\textbf{Implications.}
Our results show that standard transformers are rarely a local optimum
in the space of architectures.
Simple alternatives 
can perform much better
but sacrifice universality.
This suggests that there may be room for improved architectures that better support multiple capabilities simultaneously, such as fluency and robust reasoning.
\end{abstract}
\vspace{-5pt}

\AtBeginEnvironment{tabular}{\scriptsize}

\vspace{-5pt}
\section{Introduction}
\label{sec:introduction}
\vspace{-1pt}

\textbf{Inductive biases of transformers.}
The recent history of machine learning has seen a convergence of architectures
%uniformization of models across tasks and modalities.
across modalities.
Most state-of-the-art models for vision, language, and speech are based on transformers, barring only minor differences \citep{vaswani2017attention}.
This success contrasts with earlier task-specific models, and it
has prompted the hypothesis that transformers
implement very generic inductive biases\footnotemark\;%
well suited to various types of real-world data \citep{goldblum2023no,teney2024neural,teney2025we}.
%such as a \emph{simplicity bias} akin to Occam's razor
%--- combined a high capacity.
%to fit large datasets~\citep{}.
%The simplicity bias of neural networks depends
%on architectural choices such as their activation functions \citep{teney2024neural,teney2025we}.
%Yet, 
Considering the vast space of possible architectures, the following question remains (Q1).
%\vspace{0.5pt}
\frameQuestion{\textit{Are transformers uniquely endowed with such
%and optimal solution with such
generic inductive biases?}}
\vspace{-3.5pt}

\textbf{Uneven performance across domains.}
Transformers perform remarkably, for example as language models.
%Paradoxically, they 
But they
also fail on elementary tasks
such as learning arithmetic 
\citep{nikankin2024arithmetic}.
These failures
%demonstrate limitations of transformers and 
have motivated a plethora of alternative components such as
positional encodings
\citep{cai2025extrapolation,jelassi2024repeat}
and attention mechanisms
\citep{katharopoulos2020transformers,saratchandran2024rethinking,schlag2021linear}.
But these designs %are generally partial fixes that never get
are rarely adopted beyond toy tasks.
This suggests that
the inductive biases of standard transformers
may not be suited to domains as different as natural language and arithmetic.
%may not always be optimal,
%may not be so universal,
%and that different tasks such as natural language and arithmetic 
%could benefit from different inductive biases.
%are
%not as well suited
%This raises another question (Q2).
This raises the following  question (Q2).
%\vspace{-9pt}
\frameQuestion{\textit{Should we even seek to address
such different domains with the same learning method?}}

\footnotetext{The \textit{inductive biases} of a learning algorithm can be seen as a prior over the space of functions
\citep{mitchell1980need,mingard2021sgd} such that particular (types of) functions are favored among the many that fit the data.
We focus on biases encoded in architectures,
not in choices of optimizer, objective function, initialization, etc.}

\begin{figure}[t!]
    \centering
    \includegraphics[height=0.245\linewidth]{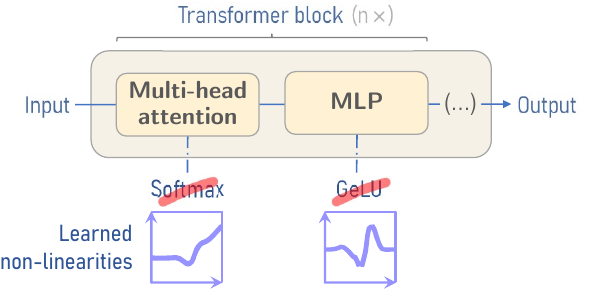}
    \hspace{8pt}
    \includegraphics[height=0.240\linewidth]{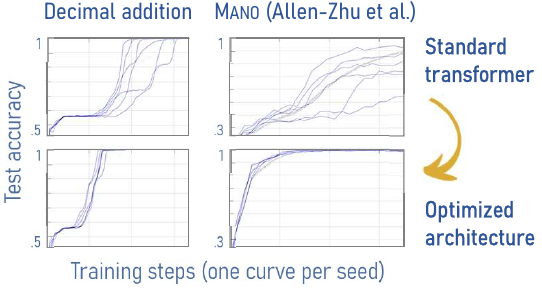}
    \vspace{-3pt}
    \caption{
    \textbf{Our approach to discover better task-specific inductive biases.}
    (Left)~We replace the non-linearities in a transformer (softmax, GeLUs)
    with splines optimized from scratch for one chosen dataset.
    (Right)~The new resulting architectures (with frozen splines as non-linearities)
    allow us to train models with 
    dramatically better convergence, generalization, and stability across seeds,
    on algorithmic tasks and code/language modeling datasets.
    We also mix-and-match the new
    architectures across tasks (not pictured)
    to evaluate the compatibility of inductive biases across tasks.
    }
    \label{fig:teaser}
    \vspace{-14pt}
\end{figure}

The above questions are relevant for developing better AI systems.
Current designs have largely relied on scaling up models and data \citep{mayilvahanan2025llms}.
This growth is not infinitely sustainable.
and their deficiencies (e.g.\ in systematic generalization and data efficiency)
indicate a need for
%that we should search for %architectures and learning methods with
better inductive biases.
Understanding the suitability of the inductive biases of transformers to specific tasks (Q1)
and the compatibility between tasks (Q2)
are steps in this direction.

\vspace{-1pt}
\textbf{Our approach.}
We address the above questions
with a method that seeks better inductive biases for a chosen dataset
by tweaking the transformer architecture.
We replace non-linearities (GeLUs, softmax) with parametrized splines,
optimized on held-out data.
This yields new architectures
that match or surpass standard transformers.
%on various %individual
%tasks.
The improvement in learning speed and/or generalization
indicates how far the standard transformer is from a local optimum in the space of architectures for a specific task (Q1).
%Furthermore, 
We also mix-and-match these new architectures across tasks
to assess how the inductive biases tuned for one task
perform for another,
thus assessing their compatibility~(Q2).

\vspace{-1pt}
\textbf{Findings.}
We study %a dozen tasks across
two domains: algorithmic skills and language modeling.
For algorithmic skills, we use toy tasks commonly used to evaluate architectures
%and motivate new models,
(see e.g.\ \citealt{allenzhu2025canon}).
For %nearly
all considered tasks,
our approach finds architectures that
dramatically improve learning speed,
%(length)
generalization, and stability across random seeds
(\autoref{sec:generalization}).
%(even out-of-distribution length generalization; \autoref{sec:generalization}).
Our task-specific variants of transformers
%can thus be
are \textbf{vastly superior to standard designs,
using only minor modifications} such as replacing the GeLUs.
The cross-task evaluation also
reveals that the new architectures are quite task-specific.
%and rarely help beyond similar tasks.
This explains why many components proposed in the literature (e.g.\ attention mechanisms, positional encodings)
%do not see adoption
are rarely adopted beyond toy tasks.
These results also challenge the view that a single architecture can be optimal for a vast set of tasks \citep{goldblum2023no}.

\vspace{-3pt}
For language modeling,
we evaluate multiple datasets
%of various sizes and complexity, covering
of natural language and computer code. %(English, Python, Java).
%, using character-level and tokenized versions.
In most cases, we find 
optimized architectures that slightly improve over a baseline transformer.
We stress that these improvements are practically not directly useful, %in practice,
because standard components are more computationally efficient.
But they matter indirectly, because they are evidence that
\mbox{\textbf{standard transformers are neither a unique nor a local optimum}}
in the space of architectures. %for any of these standard datasets.
%But they show that \textbf{vanilla transformer are not a local optimum in the space of architectures} for any of these standard datasets.
In contrast to algorithmic tasks,
the cross-task evaluation shows here that the improvements
can transfer across natural language datasets, and across tokenization levels (character vs.\ subword).
Overall, our results indicate that
standard transformers are intrinsically
better suited to modeling natural language than code, %and algorithmic skills.
%the architectures best suited
%to natural language %modeling
%are not be simultaneously optimal for code,
and clearly ill-equipped to learn algorithmic skills.

\vspace{-1pt}
%\textbf{Summary of contributions.}\vspace{2pt}
Our contributions are summarized as follows.
\begin{itemize}[leftmargin=*, itemsep=1.5pt, parsep=0pt, topsep=-4pt] % Less space
%\begin{itemize}[leftmargin=*, itemsep=3pt, parsep=0pt, topsep=-3pt] % More space
\item
\textbf{A method to optimize a transformer architecture} for any given dataset (\autoref{sec:methods}).
%as a way to approximate optimal, task-specific inductive biases. 
We replace GeLUs and softmaxes with parametrized components %based on splines
optimized on held-out data. The optimized architecture can then be used with standard training to evaluate its suitability to any other dataset.

\item
\textbf{An application to algorithmic tasks} (\autoref{sec:experimentsAlg}).
We find that optimized architectures
dramatically improve learning speed, 
generalization, and stability across seeds.
%on algorithmic tasks (pictured) and language modeling.
They also prove very task-specific, showing
%that these tasks require
the utility of
inductive biases very different from %those of
standard transformers'.

\item
\textbf{An application to language modeling} (\autoref{sec:experimentsLang}).
We obtain small, albeit consistent improvements,
showing that standard transformers are neither unique nor optimal designs,
even for common code and natural language modeling tasks.

%\item We also discuss implications for the development of future learning systems in \autoref{sec:discussion}.
%The improvement also hold while scaling the width and depth of the models.

\begin{comment}
\item
\textbf{Further analysis of models} based on standard vs.\ optimized architectures (\refApp{app:mdl}).
We find that the optimized architectures
lead to models with reduced sensitivity to weight perturbations,
which we interpret as a smaller MDL
and thus a better compression of the training data.
\end{comment}
\end{itemize}

\vspace{-1pt}
We discuss the implications for the development of future AI models
%learning systems
in~\autoref{sec:discussion}.
\clearpage
%\vspace{-5pt}

%\section{Proposed Methods to Optimize and Evaluate Architectures}
\section{Proposed Method to Optimize Architectures}
\label{sec:methods}

\textbf{Goal.}
As a baseline architecture, %for sequence modeling,
we consider
a standard decoder-only transformer
(GPT-2-style, see details in \refApp{app:implementation}).
Our goal is to evaluate whether this choice is 
optimal for specific %tasks and
datasets.
We also seek
better variants of transformers,
i.e.\ identifying inductive biases better suited to each task.
Evaluating the new architectures across tasks can then measure the compatibility of
pairs of tasks. 
All the tasks we consider are formulated as sequence completion of natural language, computer code, or abstract tokens.

\vspace{4pt}
\textbf{Replacing non-linearities with learnable parametrized functions.}
We replace the main non-linearities in a transformer 
with components that can be optimized (see \autoref{fig:teaser}).
Indeed, the main difference between a transformer and a simple linear model
hinges on a few non-linear operations in the attention and MLP layers,
which we will alter to obtain different inductive biases.
%\begin{itemize}[leftmargin=*, itemsep=2pt, parsep=0pt, topsep=-5pt] % Less space
%\begin{itemize}[leftmargin=*, itemsep=3pt, parsep=0pt, topsep=-3pt] % More space
\begin{itemize}[leftmargin=*, itemsep=5pt, parsep=0pt, topsep=-2pt] % Lots of space
\item
An MLP layer is defined as:
$\bx\!\leftarrow\!\bW' \,\phi\big(\bW \bx + \bb\big) + \bb'$
where $\bx$ is a vector of activations, $\bW$, $\bW'$, $\bb$, $\bb'$ learned weights and biases, and
$\phi\!:\!\mathbb{R}\!\to\!\mathbb{R}$
an element-wise non-linearity.
In the baseline architecture, $\phi$ is a GeLU.
In our model, 
$\phi_{\btheta_\mathrm{MLP}}$
is a 1D linear spline parametrized by 
learnable keypoints
$\btheta_\mathrm{MLP}$,
capable of approximating a variety of functions (details in \refApp{app:implementation}).
\item
An attention layer in the baseline transformer is defined as:
$\bx\!\leftarrow\! \operatorname{softmax} \! \left( \bQ \, \bK\!^\top \!\right) \, \bV$,
where
$\bx$ is the output vector of activations
and $\bQ,\bK,\!\bV$
are linear projections of the input.
This is a special case of the kernel version of attention:
$\bx\!\leftarrow\! 
{\sum_{j=1}^n K(\bQ_i, \bK_j)\,\bV_j}
\,\big/\,
{\sum_{j=1}^n K(\bQ_i, \bK_j)} \,$
where the similarity
between $\bQ$ and $\bK$ is measured with a kernel function $K(\bQ,\bK)$.
In the baseline transformer, 
%exponential dot-product kernel
$K_{\mathrm{smax}}(\bQ,\bK) \!=\! \exp\!\big(\bQ^\top \bK/\sqrt{d}\big)$. %
%\footnote{The function $K_{\mathrm{smax}}(\cdot,\cdot)$ can also be viewed as a kernel corresponding to an infinite-dimensional feature map.}
In our model,
we introduce
a learnable non-linearity
%a learnable parametrization with an element-wise non-linearity
$\phi'\!:\!\mathbb{R}\!\to\!\mathbb{R}$
giving
$K(\bQ,\bK) \!=\! \phi'(\bQ)^{\top}\phi'(\bK)$.
We implement
$\phi'$
as a linear spline
$\phi'_{\btheta_\mathrm{A}}$
with keypoints $\btheta_\mathrm{A}$
that can be optimized. %to modify the different inductive biases.
\end{itemize}\vspace{2.5pt}

\vspace{4pt}
\textbf{Two-stage setting.}
Our experiments proceed in two stages.
In stage~\rom{1}, we optimize the architecture for a chosen dataset~$\mathbb{D}$
by training both the model's weights and its parametrized non-linearities
($\btheta_\mathrm{A},\btheta_\mathrm{MLP}$)
on~$\mathbb{D}$.
In stage~\rom{2}, the non-linearities are frozen,
and we retrain the model in a standard manner from scratch on any dataset~$\mathbb{D}'$.
The models obtained from stage~\rom{2} are thus fairly comparable with the baseline architecture.%
\footnote{In stage~\rom{2}, %the non-linearities are frozen and %their parameters
($\btheta_\mathrm{A},\btheta_\mathrm{MLP}$)
are frozen and better viewed as pre-tuned~hyperparameters than extra model capacity.}
When $\mathbb{D}'\!\neq\!\mathbb{D}$,
i.e.\ a ``mix-and-match'' setting,
stage~\rom{2} serves to evaluate whether the inductive biases
optimized for~$\mathbb{D}$ suit the learning of~$\mathbb{D}'$.

%mix-and-match to study the compatibility of the optimized AFs/inductive biases across datasets

\vspace{4pt}
\textbf{Optimizing architectures.}
Our method may seem similar
to prior work about learning activation functions (e.g.\ \citep{alexandridis2025adaptive})
but their goals are very different.
These works seek to improve performance by continuously updating the activation during training.
%adding flexibility and capacity to the model.
%Our goal is very different: 
Whereas we seek to identify
inductive biases that can remain hard-encoded in the architecture and further reused
%to train multiple models.
to train new models with other seeds and datasets (stage~\rom{2}).
We make this possible with a \textbf{two-loss training}.
During stage~\rom{1}, we hold out a fraction of the training data (e.g.\ $20\%$)
that we use solely for optimizing the non-linearities,
while we optimize the weights in a standard manner on the training set.
This prevents a co-adaptation,
that could cause the non-linearities to overfit
particular weights or seed.
This is particularly important for our experiments on algorithmic toy tasks,
and even more so for improving length generalization%
\footnote{The benefit of the two-loss training is smaller for language modeling
because the models are heavily over-parametrized and never at risk of overfitting the training data.}
(\autoref{sec:lengthGeneralization}).
In this latter case, we hold out an out-of-distribution (OOD) split of data 
(see \autoref{sec:lengthGeneralization}),
such that the weights are optimized for one range of sequence lengths,
and the architecture for a different wider range.
This forces the architecture to capture an inductive bias for length generalization.
In stage~\rom{2}, the non-linearities are frozen, and the model weights are trained in a standard manner on the whole training split of the target dataset.

A second innovation to prevent the co-adaptation of weights and non-linearities in stage~\rom{1}
is \textbf{multi-model training}.
We train $M$ models in parallel (e.g.\ $M\!=\!4$)
that use different seeds
but share the non-linearities being optimized.
The resulting optimized architecture is naturally more likely
to generalize in stage~\rom{2} to other weights and datasets (see \refApp{app:multimodel}).
This also proves particularly helpful for algorithmic tasks
because the variance across seeds of the baseline architecture is often high.
We provide a complete description of our method as \autoref{alg:alg} in the appendix.

\clearpage
\textbf{Rationale for splines.}
We parametrize our non-linearities
as linear splines
because they offer
the most unbiased, tractable parametrization for an $\mathbb{R}\!\rightarrow\!\mathbb{R}$ function.
For example, a spline can represent the identity function as easily as a step function or a sine wave.
Prior work on trainable activation functions
uses e.g.\ small MLPs that enforce priors like smoothness or monotonicity \citep{apicella2021survey,greydanus2020scaling}.
Such parametrizations would struggle to capture sharp transitions like those in Figure~\ref{fig:afAlg}.
We also favor \emph{linear} splines over higher-order (e.g.\ cubic) ones
because they behave similarly while being much cheaper, as evaluated by \citet[Appendix D]{teney2025we}.

%-----------------------------------------------------------------------------------------------

\vspace{-2pt}
\section{Experiments on Algorithmic Reasoning Tasks}
\label{sec:experimentsAlg}

In this section, we apply the proposed method
to a set of tasks commonly used to evaluate 
the algorithmic skills of transformers (\autoref{tab:tasksAlg}).
These tasks seem elementary but they
are remarkably challenging for transformers,
and often used to highlight their limitations.
The tasks are all formulated as sequence completion:
each example comprises an ``{input}'' part, followed by a separator, then an ``{output}'' part.
The models are trained for next-token prediction
on the {output} part of training sequences.
Unless otherwise noted
we use
i.i.d.\ sets of
training, validation, and test data.

\textbf{Experimental setup.}
For each task $\mathbb{D}$, we first train the baseline architecture
and tune its hyperparameters
(width, depth, learning rate, batch size, etc.)
for high accuracy and fast convergence on the validation set.
We then run the proposed method (stage~\rom{1}, $M\!=\!8$)
to optimize the architecture
for $\mathbb{D}$.
We then re-train a model from scratch with the optimized architecture (stage~\rom{2}),
keeping the same hyperparameters 
(we get no further improvements by re-tuning them).
In \autoref{sec:acrossTasksAlg}, we also re-train models
on other tasks $\mathbb{D}'$
as a way to evaluate
the compatibility between $\mathbb{D}$ and $\mathbb{D}'$,
and the generality of the optimized architectures.
All results are averages over 6 random seeds.

\vspace{-4pt}
\begin{table}[ht!]
    \centering
    \caption{Algorithmic tasks used in our experiments.
    They are sized such that they
    %have similar complexity and
    require similar model capacity, except for \textsc{Mano} \citep{allenzhu2025canon} which is intrinsically more complex.}
    \vspace{-4pt}
    \label{tab:tasksAlg}
    \setlength{\tabcolsep}{2pt}
    \renewcommand{\arraystretch}{0.95}
    \begin{tabular}{p{9.5cm} c p{4.0cm}}
    \toprule
    \textbf{Task} && \textbf{Examples}\\
    \midrule
    {\textbf{\textsc{memorize}}.
    Simple memorization of a mapping
    between a two-integer key and an integer value, with all integers in $[1,\!32]$.
    Each sequence consists of the key, a separator, and the value.
    This task has no test set:
    performance is simply the training accuracy
    %memorize-32-2
    \citep{zhong2024algorithmic}.}&&
    \texttt{23 12 | 10\newline
    11 32 | 27\newline
    31 19 | 18}\\

    \midrule
    {\textbf{\textsc{parentheses}}.
    Recognition of Dyck language.
    Each sequence contains parentheses followed by a separator and a marker
    indicating whether they are balanced or not.
    %contains an equal number of opening and closing parentheses.
    Sequences lengths are in $[1,\!20]$ in the training set, and
    $[21,\!40]$ in the validation and test sets
    %parentheses-1e5-1-20-21-40-21-40-1
    \citep{zhong2024algorithmic}.}&&
    \texttt{( ) ( | <unbalanced>\newline
    ( ( ) ( ) ) | <balanced>\newline
    ) ( ) ( ) | <unbalanced>}\\

    \midrule
    {\textbf{\textsc{AddMod}}.
    Modular addition mod\,$N$, with $95\%$ of the $N^2$ examples used for training
    %modularAddition-57-0.95
    \citep{zhong2024algorithmic}. We use $N$=$97$.}&&
    \texttt{12 3 | 15\newline
    96 2 | 1}\\

    \midrule
    {\textbf{\textsc{Haystack}}.
    Needle-in-a-haystack recall.
    The model gets a sequence
    $[m_1, c_1 ... \,m_k, c_k, m_u]$
    of markers $m_k$ and values $c_k$.
    %The input ends with a marker $m_u$.
    It must search for the first occurrence of $m_u$
    and return its successor %value
    $c_u$
    %haystack-1e5-10-64
    \citep{zhong2024algorithmic}.
    We use $k \!\in\! [1,\!10]$ and $m_k, c_k \!\in\! [1,\!64]$.}&&
    \texttt{2 p 9 k 3 b 9 | k\newline
    8 a 2 b 8 | a\newline
    2 p 9 k 3 b 5 x 5 | x}\\

    \midrule
    {\textbf{\textsc{Add}}.
    Decimal addition of %equal-length
    4-digit numbers with digit-wise tokens.
    %We use 4-digit numbers.
    %decimalAddition-1e5-4-4-0
    \citep{zhong2024algorithmic}.}&&
    \texttt{1 0 0 9 + 1 0 9 2 | 2 1 0 1}\\

    \midrule
    {\textbf{\textsc{AddReversed}}.
    \textsc{Add} 
    with reversed numbers, known to be easier to learn
    %decimalAddition-1e5-4-4-1
    \citep{lee2023teaching}
    .}&&
    \texttt{9 0 0 1 + 2 9 0 1 | 1 0 1 2}\\

    \midrule
    {\textbf{\textsc{Copy}}.
    Repeating the input. Elementary but unsolved for length generalization
    %copy-5e6-8-2-10-2-15-18-18-1
    % Hard-Alibi
    %\citep{jelassi2024repeat}
    % 2025 Extrapolation by Association - Length Generalization Transfer in Transformers
    %\citep{cai2025extrapolation}
    %2025 Self-Improving Transformers Overcome Easy-to-Hard and Length Generalization Challenges
    %\citep{lee2025self}
    \citep{cai2025extrapolation}. %,jelassi2024repeat}.
    Tokens in $[1,\!8]$.
    Seq.\ lengths in $[2,\!10]$ for training,
    $[2,\!15]$ for validation,
    $[16,\!20]$ for testing.}&&
    \texttt{2 8 | 2 8\newline
    %7 7 3 | 7 7 3\newline
    9 4 8 7 8 3 | 9 4 8 7 8 3}\\

    \midrule
    {\textbf{\textsc{Mano}}.
    %Knowledge manipulation
    Synthetic task proposed by
    \citet{allenzhu2025canon}
    to evaluate large pretrained models. %LLMs. %large-scale models and pretrained LLMs.
    Each sequence specifies nested arithmetic operations mod\,$N$
    with number-level tokens.
    Our scaled-down version uses
    $N$=$7$ and a number of operations per sequences in $[1,\!3]$.
    %trSetSize, nOperationsMin, nOperationsMax, modulus, includeTokenNOps)
    %%polMano-1e5-1-3-7-0
    }&&
    \texttt{%
    (1*3)+4 | 0\newline
    (2-(6-1))*3 | 5\newline
    (3*(5-6))-1 | 3}\\[0.55pt]
    \bottomrule
    \end{tabular}
\end{table}

\vspace{-4pt}
\begin{figure}[h!]
    \centering
    \setlength{\tabcolsep}{4pt}
    \renewcommand{\arraystretch}{1.0}
    \begin{tabular}{r cc rr cc}
    ~ & Baseline & Ours & ~~~~~ & ~ & Baseline & Ours\\%[3pt]

    \raisebox{23pt}{\textsc{Haystack}\!\!\!}&
\includegraphics[height=0.12\linewidth]{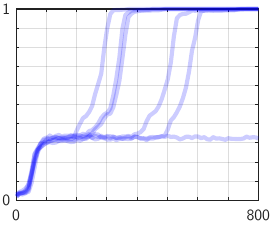}&
\includegraphics[height=0.12\linewidth]{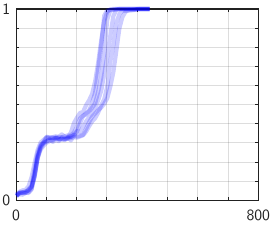}&~&

    \raisebox{23pt}{\textsc{Copy}\!\!\!}&
\includegraphics[height=0.12\linewidth]{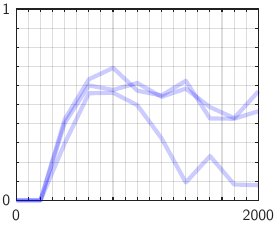}&
\includegraphics[height=0.12\linewidth]{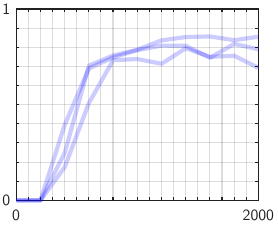}\\

    \raisebox{23pt}{\textsc{AddReversed}\!\!\!}&
\includegraphics[height=0.12\linewidth]{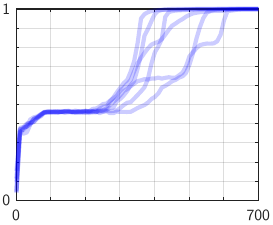}&
\includegraphics[height=0.12\linewidth]{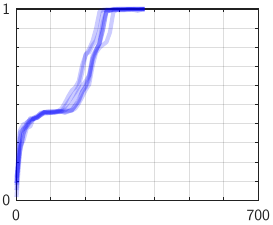}&~&

    \raisebox{23pt}{\textsc{Mano}\!\!\!}&
\includegraphics[height=0.12\linewidth]{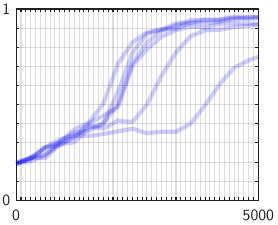}&
\includegraphics[height=0.12\linewidth]{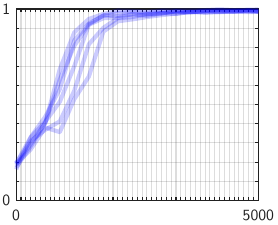}
    \end{tabular}
    \vspace{-8pt}
    \caption{Training curves (test accuracy vs.\ training step, one curve per random seed) of models trained on algorithmic tasks
    with a baseline transformer or our optimized architectures.
    The latter converge much faster and show less variance across seeds.
    See \refApp{app:trCurvesAlg} for other tasks.}
    \label{fig:trCurvesAlgSelection}
    \vspace{-16pt}
\end{figure}
%\clearpage

%----------------------------------------------------------------------------------------------
\subsection{Improvements on Individual Tasks}
\label{sec:learningSpeed}
\label{sec:generalization}
\vspace{-1pt}

%----------------------------------------------------------------------------------------------
\textbf{Faster convergence.}
The most striking improvement %when training
with optimized architectures
is the learning speed
%much smaller number of steps
%vs.\ a baseline transformer is the number of steps
%before converge. %(saturation of training and test accuracy).
(\autoref{fig:trCurvesAlgSelection}).
For the \textsc{Add} and \textsc{Mano} tasks for example, convergence occurs $2$\,--$3\times$ faster.
The learning rate of the baseline was tuned to its maximum stable value for every task.

%----------------------------------------------------------------------------------------------
\textbf{Reduced variance.}
On some tasks, baseline transformers
show huge variance in accuracy and training speed across random seeds.
This suggests tasks that are underspecified \citep{teney2021evading,teney2022predicting}
and misaligned with the model's inductive biases \citep{zhou2024transformers}.
In these cases, the optimized architectures %essentially
eliminate the problem and make the training much more reliable
(\autoref{fig:trCurvesAlgSelection}).

%----------------------------------------------------------------------------------------------
\textbf{Better generalization.}
For some tasks,
baseline transformers do not reach perfect test accuracy
though they perfectly fit the training data.
This shows again a misalignment between the target function and the inductive biases.
Optimized architectures solve this problem (see e.g.\,\textsc{Mano}, \autoref{fig:trCurvesAlgSelection}).

\begin{comment}
\begin{table}[ht!]
    \centering
    \caption{Table of results? Bar plot?
    gelu/interp_smax/gelu;interp/interp_interp
    with variance across seeds}
    \label{tab:resultsAlg}
    \vspace{-3pt}
    \setlength{\tabcolsep}{2.2pt}
    \renewcommand{\arraystretch}{1.0}
    \begin{tabular}{l ccc | cc | ccccccccc}
    \toprule
    Todo.\\
    \bottomrule
    \end{tabular}
    \vspace{6pt}
\end{table}
\end{comment}

%Even when the performance of the optimized architecture is not clearly better than the baseline, the fact that we discover different activation functions that work just as well as GeLUs shows that the latter are not a sole local optimum in the space of architectures.

%----------------------------------------------------------------------------------------------
\textbf{Improved length generalization.}
\label{sec:lengthGeneralization}
An outstanding challenge for transformers
is the generalization to sequences longer than seen during training.
Even the \textsc{Copy} task is unsolved
and a baseline transformer completely fails on unseen lengths (\autoref{fig:lengthGeneralization}).
Among the plethora of existing partial solutions,
the Alibi positional encodings \citep{press2021train}
bring non-trivial accuracy on slightly longer sequences. %($\sim\!60\%$).
We use our method to optimize the Alibi architecture.
We use the two-loss mechanism of \autoref{alg:alg}
to optimize the transformer weights on lengths $2$--$10$
and the non-linearities on $2$--$15$. This forces the optimized architecture to capture an inductive bias for length generalization.
As a result,
a model trained with the optimized architecture reaches higher accuracies
on longer sequences.
While this is not a complete solution to length generalization,
it shows that inappropriate inductive biases in the base architecture
are one of the obstacles to length generalization.

\newcommand{\lnn}[3]{%
    {\scriptsize\textcolor[HTML]{#1}{\raisebox{-0.04em}{\scalebox{1.2}[1]{\ding{110}}}}%
    \textcolor[HTML]{#2}{{#3}}}
}

\vspace{-4pt}
\begin{figure}[h!]
    \centering
    \setlength{\tabcolsep}{2pt}
    \renewcommand{\arraystretch}{1.0}
    %\hspace{-4pt}
    \begin{tabular}{rcc}
        \raisebox{39pt}{\makecell[r]{Test accuracy\\[0.2pt](sequence-wise, in \%,\\shading shows +/-$1$ std.\ dev.)}} &
        \includegraphics[height=0.18\linewidth]{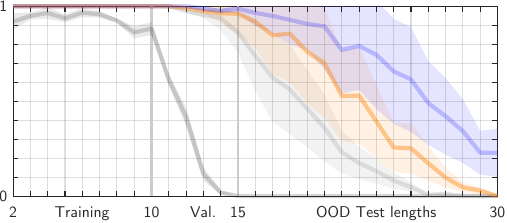}&
        \hspace{8pt}
        \raisebox{37pt}{
            \renewcommand{\arraystretch}{1.4}
            \begin{tabular}{l}
            \lnn{999999}{707070}{~Baseline}\\
            \lnn{BFBFBF}{BFBFBF}{~Alibi \citep{press2021train}}\\
            \lnn{FFC489}{BF9459}{~Alibi\,+\,Ours w/o two losses}\\
            \lnn{BFBFFF}{9F9FDF}{~Alibi\,+\,Ours (full method)}\\
            \end{tabular}
        }\\[1pt]
        ~ & Sequence length %(trained on $2$--$10$) & ~
    \end{tabular}
    \vspace{-4pt}
    \caption{Length generalization on the \textsc{Copy} task.
    The baseline completely fails on unseen lengths~($\gg\!\!10$).
    Alibi positional encodings \citep{press2021train} help.
    Optimizing the Alibi architecture with our method further improves the accuracy
    and extends the benefits to longer sequences.\liang{I know the difference and the importance of comparison between the two `ours', but I am not sure it would be super straightforward to the reader. Still thinking whether/how to improve.}}
    \label{fig:lengthGeneralization}
    %\vspace{6pt}
\end{figure}

\vspace{-6pt}

%----------------------------------------------------------------------------------------------
\textbf{Performance with smaller models.}
We train models of different widths for each task.
%using either the baseline or optimized architecture.
Results in \autoref{fig:algCapacity} %in the appendix
show that the accuracy drops more sharply on some tasks with the baseline architecture
than optimized ones.
Intuitively, when the architecture is already aligned with the task,
less capacity is needed in its weights.
Equivalently, a fixed number of parameters offers more capacity.
%to fit particular data points.

% Colored thick line, different color for the text than the square
\newcommand{\lnnn}[1]{%
    {\textcolor[HTML]{#1}{\raisebox{-0.04em}{\scalebox{1.05}[1]{\ding{110}}}}}\!
}

\begin{figure}[h!]
    \vspace{-4pt}
    \centering
    \setlength{\tabcolsep}{1pt}
    \renewcommand{\arraystretch}{1.0}
    \begin{tabular}{r ccc c ccc}
    &
    \multicolumn{3}{c}{\scriptsize Better accuracy on small models}& \hspace{10pt} &
    \multicolumn{3}{c}{\scriptsize No clear difference}\\
    \cmidrule{2-4}
    \cmidrule{6-8}\\[-8pt]
    \raisebox{9pt}{\rotatebox{90}{\scriptsize Test accuracy (\%)}}&
    \includegraphics[height=0.145\linewidth]{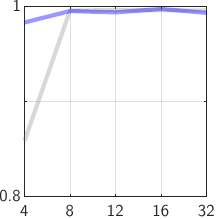}&
    \includegraphics[height=0.145\linewidth]{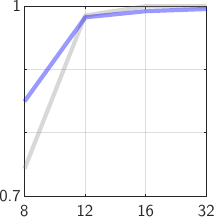}&
    \includegraphics[height=0.145\linewidth]{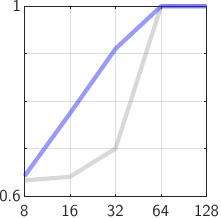}&&
    \includegraphics[height=0.145\linewidth]{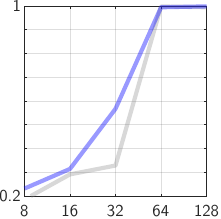}&
    \includegraphics[height=0.145\linewidth]{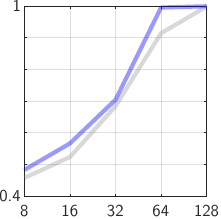}&
    \includegraphics[height=0.145\linewidth]{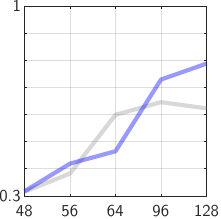}\\[2.5pt]
    &\textsc{parentheses}&
    \textsc{memorize}&
    \textsc{AddReversed}&&
    \textsc{Haystack}&
    \textsc{Add}&
    \textsc{AddMod}
    \end{tabular}
    \vspace{-5pt}
    \caption{Test accuracy of models of different widths ({\small\textsc{X}}\,axis).
    On some tasks, %models with the
    optimized architectures (\lnnn{9F9FFF})
    maintain higher accuracy than the baseline (\lnnn{AAAAAA})
    when reducing the width of the model.}
    \label{fig:algCapacity}
    %\vspace{3pt}
    %\vspace{-120pt}
\end{figure}

\begin{comment}
little benefits in optimizing the attention:
not a strong conclusion
maybe just because of our limited search space
we tried a less constrained parametrization of the attention mechanism, but was difficult to optimize
(failing to reach the performance of a standard softmax attention)
\end{comment}

%----------------------------------------------------------------------------------------------
\clearpage
\subsection{Compatibility of Optimized Architectures Across Algorithmic Tasks}
\label{sec:acrossTasksAlg}

\begin{figure}[h!]
    \centering
    \hspace{-49pt}
    \begin{tabular}{rc}
    \raisebox{60pt}{\makecell[r]{\scriptsize Architectures\\
        \scriptsize optimized for\\
        \scriptsize specific tasks}}&
        \includegraphics[height=0.345\linewidth]{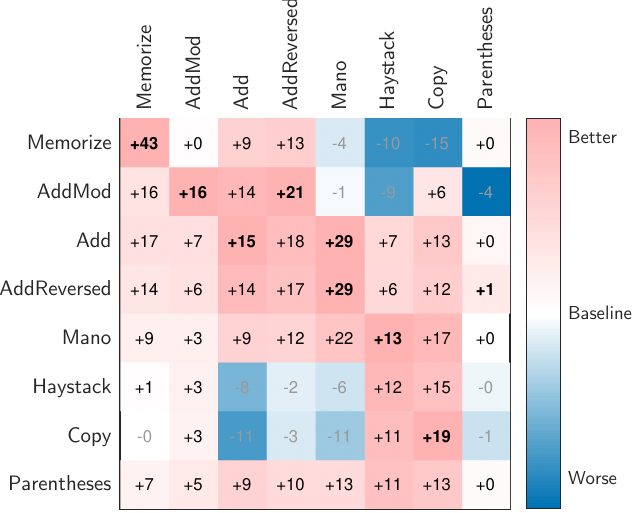}\\[2pt]
    ~ & \scriptsize Target tasks
    \end{tabular}
    \vspace{-4pt}
    \caption{Compatibility of architectures across algorithmic tasks. 
    We plot the absolute difference in test accuracy (\%) with the baseline after a fixed number of steps
    %to captures improvements in both generalization and training speed
    (details in \refApp{app:implementation}).
    %See text for details.
    The best option %architecture
    per task (column)
    is usually on the diagonal,
    meaning that the optimized architectures are quite task-specific,
    while still yielding some positive transfer.}
    \label{fig:acrossTasksAlg}
\end{figure}

We now train models on each task $\mathbb{D}$
using %the best
architectures optimized for any other task $\mathbb{D}'$
to evaluate the pairwise compatibility of their inductive biases.
The results in \autoref{fig:acrossTasksAlg}
show that the optimized architectures are very task-specific.
Few of the benefits transfer across tasks, mostly across closely related tasks
like \textsc{Add} and \textsc{AddReversed}.
Many perform worse than a standard transformer.
This shows that the specialization to our algorithmic tasks comes at the cost of universality.
These tasks are very narrow however
and it remains an open question
whether the negative impact %on other tasks
is inevitable.
A future step to study this question could be a multi-task optimization in \autoref{alg:alg}.

\vspace{-1pt}

\vspace{-2pt}

\begin{figure}[h!]
    \centering
    \setlength{\tabcolsep}{3pt}
    \renewcommand{\arraystretch}{1.0}
    %\hspace{-4pt}
    \begin{tabular}{cccccccc}
        \includegraphics[trim=0pt 137pt 0pt 0pt,clip,width=0.119\linewidth]{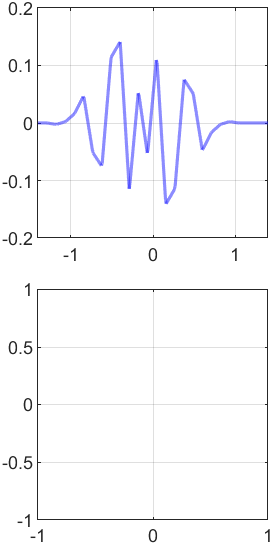}&
        \includegraphics[trim=20pt 137pt 0pt 0pt,clip,width=0.105\linewidth]{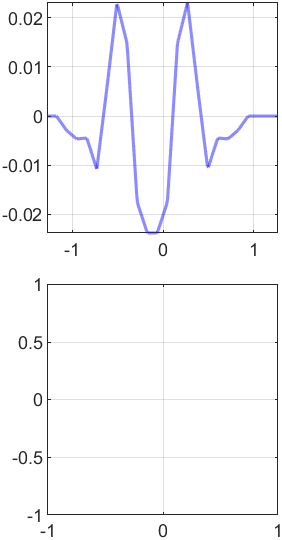}&
        \includegraphics[trim=15pt 137pt 0pt 0pt,clip,width=0.105\linewidth]{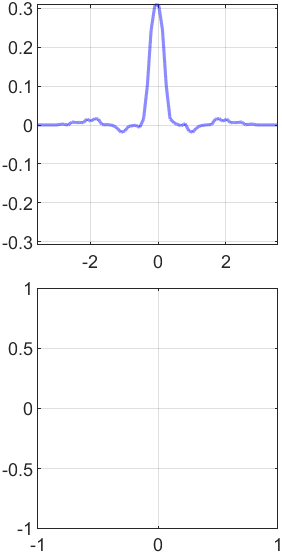}&
        \includegraphics[trim=25pt 137pt 0pt 0pt,clip,width=0.105\linewidth]{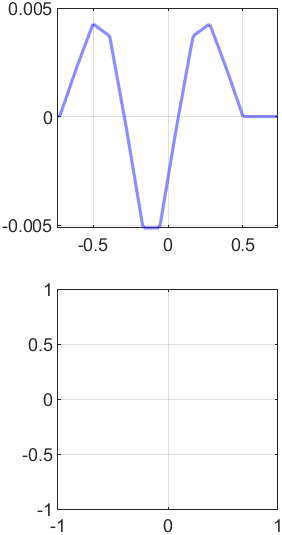}&
        \includegraphics[trim=20pt 137pt 0pt 0pt,clip,width=0.105\linewidth]{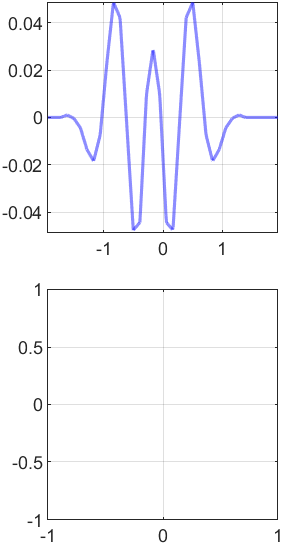}&
        \includegraphics[trim=20pt 137pt 0pt 0pt,clip,width=0.105\linewidth]{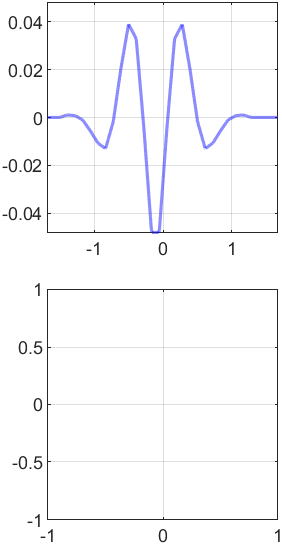}&
        \includegraphics[trim=20pt 137pt 0pt 0pt,clip,width=0.105\linewidth]{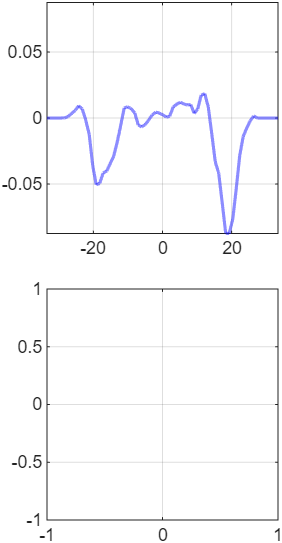}&
        \includegraphics[trim=20pt 137pt 0pt 0pt,clip,width=0.105\linewidth]{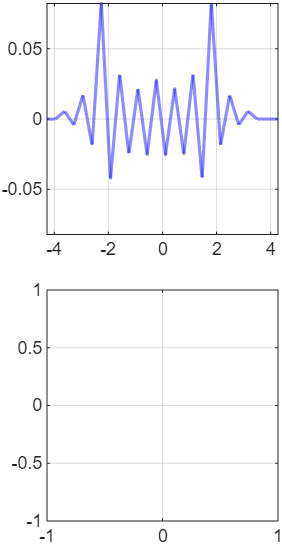}\\[3pt]
        ~~~\textsc{memorize}&
        \textsc{parentheses}&
        \textsc{AddMod}&
        \textsc{Haystack}&
        \textsc{Add}&
        \textsc{AddReversed}&
        \textsc{Copy}&
        \textsc{Mano}
    \end{tabular}
    \vspace{-3pt}
    \caption{MLP non-linearities optimized for each algorithmic task.\liang{This is not referred/analyzed in the text if I see correctly}}
    \label{fig:afAlg}
    %\vspace{6pt}
\end{figure}

\vspace{-4pt}

\myFrame{\textbf{Take-away.}~~On algorithmic tasks,
optimized architectures can
dramatically outperform standard transformers,
but the benefits are quite task-specific.
This means that these tasks %represent skills that
require inductive biases very different
from those of standard transformers.}
\vspace{2pt}

%\clearpage

\section{Experiments on Language Modeling}
\label{sec:experimentsLang}

We now apply the same experimental setup as \autoref{sec:experimentsAlg}
to language modeling.
We use datasets for
computer code (English, Java)
and natural language of various complexity levels
(\autoref{tab:tasksLang}).
Our goal is to understand whether different types of data
benefit from different inductive biases.
Current practices for building LLMs
show that data diversity is beneficial \citep{longpre2024pretrainer}
% goyal2024scaling: Scaling Laws for Data Filtering--Data Curation cannot be Compute Agnostic
and that code is complementary to natural language
\citep{aryabumi2024code,petty2024does}.
But because 
all kinds of data are mixed during training,
it is unknown whether
they could each exploit or elicit different
mechanisms in a model.
We also consider versions of the datasets tokenized
%tokenization
at the character or subword level
(BPE; details in \refApp{app:implementation}).
%(using BPE, see \citet{gage1994new} and \refApp{app:implementation}).
These choices are motivated by \citet{mayilvahanan2025llms}
who showed that 
LLM performance
is mostly determined by
data diversity and tokenization.
%the nature of the training data and its tokenization
%are the most important factors for LLM performance.

\begin{table}[ht!]
    \centering
    \caption{Datasets used in our experiments for language modeling
    (see \refApp{app:implementation} for details).}
    \vspace{-4pt}
    \label{tab:tasksLang}
    \setlength{\tabcolsep}{2pt}
    \renewcommand{\arraystretch}{1.0}
    \begin{tabular}{p{9.5cm} c p{4.0cm}}
    \toprule
    \textbf{Dataset} && \textbf{Excerpt}\\

    \midrule
    {\textbf{\textsc{TinyStories}.}
    Children stories generated with GPT-3.5.
    It was designed to capture core aspects of natural language
    (syntax, coherence, compositionality)
    with a limited vocabulary.
    This allows smaller-scale experiments than web-scale open-domain corpora
    %without requiring massive compute
    %It contains grammatically and semantically correct natural language (English)
    %while remaining much more predictable than web-scale open-domain corpora.
    %It is widely used for training and evaluating small experimental LLM architectures because it provides a clean, controlled benchmark that captures core aspects of language modeling (syntax, coherence, compositionality) without requiring massive compute or data, making it ideal for rapid iteration and testing of new model designs.
    \citep{eldan2023tinystories}.}&&
    \texttt{Once upon a time, there was a clever little dog named Max. Max loved to run
    %and play with his friends
    %in the park. One day, Max was running very fast when he fell and hurt his knee. Max went to his friend, the wise old owl, and said, "Owl, my knee hurts. What can I do?" The owl thought for a moment and said, "Max, you should test your knee. Try to walk slowly and see if it still hurts."
    } (...)\\
    \midrule
    {\textbf{\textsc{Shakespeare}.}
    Plays and sonnets by William Shakespeare, often used in early research on language modeling. It includes recognizable patterns of grammar, rhythm, and vocabulary, as well as a unique structure because of the speaker labels and dialogue formatting
    %It was a popular benchmark for RNN-based models since it is compact, easy to process, and highlights a model's ability to capture long-range dependencies and stylistic regularities without requiring large-scale data.
    \citep{karpathy2015charRNN}.}&&
    \texttt{BENVOLIO: Good-morrow, cousin.\ ROMEO: Is the day so young?\ BENVOLIO: But %new struck nine.
    %ROMEO:
    %Ay me! sad hours seem long.
    %Was that my father that went hence so fast?
    } (...)\\

    \midrule
    {\textbf{\textsc{enwik8}.}
    First 100\,M %$10^8$
    bytes of the English Wikipedia
    %typically used as a text compression benchmark
    \citep{enwik8}.
    We use the clean version from \citet{clean8}
    with only text visible to human readers, without links and meta data.
    This data provides dense, real-world text with a mix of vocabulary, syntax, and formatting.
    }&&
    \texttt{anarchism originated as a term of abuse first used against early working
    %class radicals including the diggers of the english revolution and the sans culottes of the french revolution
    } (...)\\

    \midrule
    {\textbf{\textsc{CodeSearchNet-Java} \& \textsc{-Python}.}
    %Benchmark
    Dataset of computer code originally created to support research on code search and code–text understanding \citep{husain2019codesearchnet}.
    We discard comments and descriptions in natural language following
    \citet{lu2021codexglue}
    to focus exclusively on code.
    %This data provides structured, domain-specific data.
    }&&
    %def sorted(self, iterator, key=None, reverse=False):
    %global MemoryBytesSpilled, DiskBytesSpilled
    \texttt{batch, limit = 100, self.\_next\_limit()
    \newline
    %chunks, current_chunk = [], []
    it = iter(it)
    } (...)\\[0.55pt]
    \bottomrule
    \end{tabular}
    \vspace{5pt}
\end{table}

%----------------------------------------------------------------------------------------------
\subsection{Improvements on Individual Datasets}

\begin{figure}[t!]
    \vspace{-9pt}
    \centering
    \setlength{\tabcolsep}{2pt}
    \renewcommand{\arraystretch}{1.0}
    \begin{tabular}{rc}
    ~ & \hspace{-19pt}Transformer width\\[3pt]
     \raisebox{29.3pt}{\rotatebox{0}{Num.\ layers}~} &
     \includegraphics[height=0.1355\linewidth]{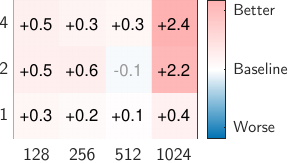}\\
    \end{tabular}
    \hspace{18pt}
    \setlength{\tabcolsep}{10pt}
    \raisebox{2.1pt}{
    \begin{tabular}{cc}
    \hspace{0pt}Optimized from scratch &
    \hspace{0pt}Optimized from a GeLU \\[3pt]
        \includegraphics[trim=0pt 137pt 0pt 0pt,clip,height=0.129\linewidth]{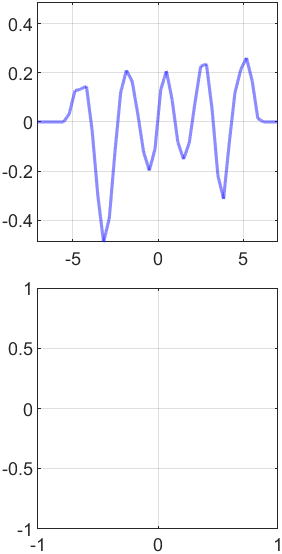}&
        \includegraphics[trim=0pt 137pt 0pt 0pt,clip,height=0.129\linewidth]{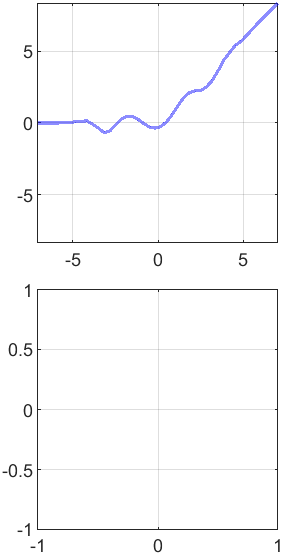}
    \end{tabular}}\hspace{15pt}
    \vspace{-2pt}
    \caption{%
    \textbf{(Left)}~Absolute improvements in token prediction accuracy (\%)
    of the best optimized architectures %over a baseline transformer
    on \textsc{TinyStories} compared to our baseline transformer.
    The accuracy is consistently slightly better at different model sizes.
    %, and the improvement is larger with wider and deeper models.
    \textbf{(Right)}~Visualization of MLP non-linearities optimized 
    from scratch (results on the left) or from a GeLU initialization (\textit{GeLU\,+\,Ours} in \autoref{fig:allLang}).
    Although %these functions
    they resemble generic wavelets, we show in \refApp{app:cleanAfs} that fine details in these functions matter.}
    \label{fig:tinyStories}
    \vspace{-10pt}
\end{figure}

\textbf{\textsc{TinyStories}.}
We compare
in \autoref{fig:tinyStories}
models trained with baseline or optimized architectures.
%at different sizes
The latter do slightly better.
The improvement is small but consistent at different model sizes.
Training curves (\autoref{fig:trCurvesLang})
show that the improvement %advantage of an optimized architecture
is larger early during training then diminishes.
%\newpage
%For the optimized architectures, 
We find it best
to optimize non-linearities only in MLPs (i.e.\ replacing GeLUs; see \autoref{fig:allLang}).
Replacing softmaxes with learned components barely matches or underperforms the baseline,
indicating a difficult optimization. %problem.
%landscape for the attention non-linearities.
%and a search space
%for our learnable attention component that is ill-suited.
%We experimented with alternative parametrizations and less restrictive formulations,
%but this made the optimization even more difficult.
We experimented with alternative parametrizations that exactly mimic a softmax at initialization.
This solution would barely move away from this initialization (not reported in tables),
suggesting that a softmax is close to a local optimum.

We visualize in \autoref{fig:tinyStories}~(right)
the optimized MLP non-linearities,
which are remarkably similar to sine wavelets.
We evaluate a non-exhaustive selection of 
activation functions
and attention variants from the literature in \autoref{tab:tinyStories}.
None of them works better than ours.
The gated linear units (\textsc{glu}s) are a popular design that adds multiplicative interactions to the MLPs.
We show that we can also improve them by introducing our learned spline in \textsc{glu}s
in lieu of their internal Swish activations.
This provides similar improvements as over standard MLPs, cf.\ \textit{\textsc{glu}/Swish} and \textit{\textsc{glu}/Ours} 
in \autoref{tab:tinyStories}.
We also evaluate
in \refApp{app:cleanAfs}
the importance of fine details
in the learned non-linearities.
We try to make them more periodic or symmetric,
but they then always perform worse.

\definecolor{darkGray}{rgb}{0.6,0.6,0.6}
\newcommand{\gelu}{\color{darkGray}GeLU}
\newcommand{\smax}{\color{darkGray}smax}
\newcommand{\rot}[1]{\rotatebox{0}{#1}}

% linear;smax & gelu;smax & gelu+interp;smax & swish+glu;smax & interp+glu;smax & relu;smax & relu2;smax & tanh;smax & sinc;smax & gaussian;smax & gelu;pOne+divSeq & gelu;pThree+divSeq & gelu;adaptive & gelu;normsmax
% 1.78 & 1.58 & \textbf{1.57} & 1.59 & \underline{1.58} & 1.60 & 1.60 & 1.71 & 2.50 & 1.64 & 1.62 & 1.60 & 1.58 & 1.58 & 
% 59.9 & 63.7 & \textbf{64.4} & 63.7 & \underline{64.0} & 63.5 & 63.6 & 61.2 & 47.7 & 62.8 & 63.0 & 63.7 & 63.7 & 63.7 & 

\begin{table}[ht!]
    \vspace{-3pt}
    \centering
    \caption{Performance of models trained on \textsc{TinyStories}
    with existing alternative attention and MLP designs 
    ($2$ layers, width $256$). None works better than ours. See \refApp{app:alternatives} for references.}
    \label{tab:tinyStories}
    \vspace{-3pt}
    \setlength{\tabcolsep}{2.2pt}
    \renewcommand{\arraystretch}{1.0}
    \begin{tabular}{l ccc | cc | ccccccccc}
    \toprule
    \textbf{Attention}
      &
      \rot{\smax} &
      \rot{\smax} & \rot{\smax} &
      \rot{\smax} & \rot{\smax} &
      \rot{\smax} & \rot{\smax} & \rot{\smax} & \rot{\smax} & \rot{\smax} & \rot{P1} & \rot{P3} & \rot{Adaptive} & \rot{NormSmax}\\
    \textbf{MLP}
      &
      \rot{Linear} &
      \rot{\gelu} & \rot{Ours} &
      \rot{\textsc{glu/}Swish} & \!\!\!\rot{\textsc{glu/}Ours} &
      \rot{ReLU} & \rot{ReLU$^2$} & \rot{TanH} & \rot{Sinc} & \rot{Gaussian} & \rot{\gelu} & \rot{\gelu} & \rot{\gelu} & \rot{\gelu}\\
    \midrule
    Tr.\ perplexity &
    1.78 & 1.58 & \textbf{1.57} & 1.59 & \!\!\!\underline{1.58} & 1.60 & 1.60 & 1.71 & 2.50 & 1.64 & 1.62 & 1.60 & 1.58 & 1.58 \\
    Val.\ acc.\ ($\%$) &
    59.9 & 63.7 & \textbf{64.4} & 63.7 & \!\!\!\underline{64.0} & 63.5 & 63.6 & 61.2 & 47.7 & 62.8 & 63.0 & 63.7 & 63.7 & 63.7 \\
    \bottomrule
    \end{tabular}
    \vspace{-3pt}
\end{table}

%While they may look nicer, they almost always perform worse than the original ones.
%but always 
%to enforce symmetry or better periodicity then retrain models with these designs.
%Although these non-linearities look somewhat nicer, the models always perform worse than the exact optimized version.

%----------------------------------------------------------------------------------------------
%\newpage
\vspace{10pt}
\textbf{\textsc{Shakespeare} \& \textsc{enwik8}.}
These datasets differ from TinyStories
in their richer vocabulary and sentence structure.
\textsc{Shakespeare} also follows a particular formatting
presenting dialogues with speaker labels (see \autoref{tab:tasksLang}).
The results in Figures~\ref{fig:allLang}\,\&\,\ref{fig:shakespeareChar}
%fig:shakespeareChar
%fig:shakespeareSubword
show that \textit{some} optimized architectures
slightly improve over the baseline.
Optimizing non-linearities in the MLPs is again more useful than in the attention.
However, differences with the baseline are small,
which suggests that standard transformers are inherently well-suited to language modeling.

The improvement is slightly clearer on \textbf{character-level datasets} than on tokenized ones (marked \textsc{-char} in \autoref{fig:allLang}).
We hypothesize that the target function to be learned by the transformer layers
for character-level language modeling
is more complex, because of the lesser capacity available in the model's token embeddings (embeddings can otherwise make up a significant fraction of the model parameters for tokenized datasets).
This could be the reason why learned non-linearities are particularly helpful, since they can help learn and represent complex functions \citep{teney2025we}.
%(i.e.\ overcoming the simplicity bias of the baseline architecture).

\begin{comment}
On the subword versions of \textsc{Shakespeare} and \textsc{enwik8},
the best optimized architectures barely reach the performance of the baseline.
Character-level language modeling is intrinsically more difficult
and requires larger models to reach a given level of performance.
Since we evaluate models of equal size for both tokenization levels,
%all datasets
%in \autoref{fig:allLang},
the models for character-level tasks can be considered
smaller relative to the difficulty of the task (as also seen from their higher training loss).
This means that the optimized architectures are more useful for small models.
It agrees with the conventional belief that
choices of architecture become less important with scale
\citep{bachmann2023scaling,tay2022scaling}.
This also means that improving the inductive biases of language models
can be key for scaling them down.
\end{comment}

We also evaluate a version of our optimized \textbf{MLP non-linearities
initialized as a GeLU} rather than a constant zero
(\textit{GeLU\,+\,Ours} in \autoref{fig:allLang}).
With this, the model starts stage~\rom{1} with
a non-linearity known to perform well. And because the optimization is non-convex,
the optimized solution remains in the local search space near GeLUs (see \autoref{fig:tinyStories}, right).
The models trained with these non-linearities
perform in-between GeLUs and those optimized from scratch.
This means that GeLUs are usually not an optimal solution, not even a local one.
But note also that our best solutions are not guaranteed to be \emph{globally} optimal
and better ones may exist.
%yet to be discovered.

%tay2022scaling,
%Scaling laws vs model architectures: How does inductive bias influence scaling?  
%bachmann2023scaling,
%Scaling mlps: A tale of inductive bias

% Colored square
\newcommand{\sq}[1]{%
    {\scriptsize\textcolor[HTML]{#1}{\raisebox{-0.04em}{\scalebox{1.2}[1]{\ding{110}}}}}
}

\begin{figure}[t!]
    \centering
    \setlength{\tabcolsep}{2pt}
    \begin{tabular}{r ccccccc}
       &\textsc{TinyStories} &
       \textsc{Shakespeare-char} &
       \textsc{Shakespeare} &
       \textsc{enwik8-char} &
       \textsc{enwik8} &
       \textsc{Csn-Java} &
       \textsc{Csn-Python}\\[3pt]
       \raisebox{7pt}{\rotatebox{90}{\tiny Tr.\ perplexity}\hspace{-2pt}}&
       \raisebox{0pt}{\includegraphics[width=0.125\linewidth]{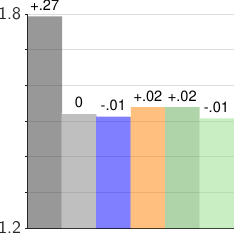}}&
       \raisebox{0pt}{\includegraphics[width=0.125\linewidth]{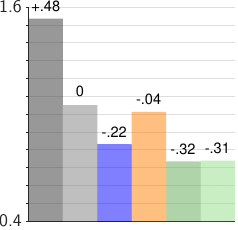}}&
       \raisebox{0pt}{\includegraphics[width=0.125\linewidth]{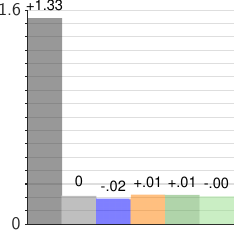}}&
       \raisebox{0pt}{\includegraphics[width=0.125\linewidth]{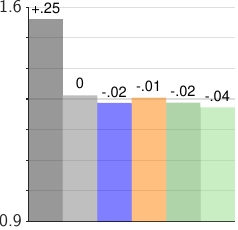}}&
       \raisebox{0pt}{\includegraphics[width=0.125\linewidth]{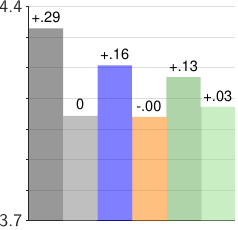}}&
       \raisebox{0pt}{\includegraphics[width=0.125\linewidth]{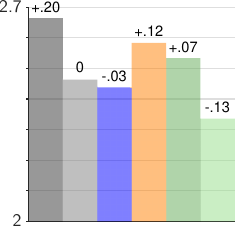}}&
       \raisebox{0pt}{\includegraphics[width=0.125\linewidth]{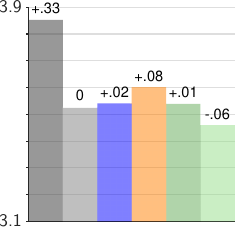}}\\[2pt]
    \end{tabular}
    \setlength{\tabcolsep}{1pt}
    \begin{tabular}{r rl rl rl rl rl rl}
     %\midrule
     \scriptsize\textbf{Attention:} &
     \multirow{2}{17pt}{\hfill \sq{999999}} & \scriptsize \smax &
     \multirow{2}{17pt}{\hfill \sq{BFBFBF}} & \scriptsize \smax &
     \multirow{2}{17pt}{\hfill \sq{7F7FFF}} & \scriptsize \smax &
     \multirow{2}{17pt}{\hfill \sq{FFBF7F}} & \scriptsize Ours & 
     \multirow{2}{17pt}{\hfill \sq{B1D5AB}} & Ours &
     \multirow{2}{17pt}{\hfill \sq{CAEEC4}} & \scriptsize \smax \\
     \scriptsize\textbf{MLP:} &&
     \scriptsize Linear &&
     \scriptsize \gelu &&
     \scriptsize GeLU\,+\,Ours &&
     \scriptsize \gelu &&
     \scriptsize Ours &&
     \scriptsize Ours
     %\midrule
    \end{tabular}
    \vspace{-6pt}
    \caption{Perplexity on code and natural language
    (lower is better; numbers on bars correspond to the difference with the baseline architecture). %datasets.
    Some optimized architectures perform slightly better than the baseline,
    often simply with optimized MLP non-linearities~(\raisebox{.04em}{\sq{CAEEC4}}\!).
    Datasets of code (\textsc{Csn-Java}, \textsc{Csn-Python}) also benefit relatively more than datasets of natural language.}
    \label{fig:allLang}
    \vspace{-10pt}
\end{figure}

%----------------------------------------------------------------------------------------------
\vspace{4pt}
\textbf{\textsc{CodeSearchNet} (\textsc{Csn-Java}, \textsc{Csn-Python}).}
The results in \autoref{fig:allLang} show
that our optimized non-linearities in MLPs
improve again over the baseline.
The gains are larger for code than for natural language,
relative to the gap between the baselines with linear and GeLU MLPs.
%(cf.\ \textit{MLP:\,Linear} and \textit{MLP:\,GeLU} in \autoref{fig:allLang}).
%This could indicate that standard transformer are even less suited to code.
%We hypothesize that this could be due to 
These larger gains may reflect the larger importance of systematic structure and compositionality in code than natural language.
The task of modeling code may thus resemble some of the algorithmic tasks of \autoref{sec:experimentsAlg},
which benefited greatly from optimized architectures.
Therefore, the architectures best suited to natural language may not be simultaneously optimal for code.

%similar trends as with other datasets.
%We slightly improve %over the baseline architecture
%by optimizing the MLP non-linearities.
%Even when differences across architectures are marginal,
%the fact that we discover multiple solutions that perform equally well
%means that the original GeLUs are not a uniquely optimal choice.
%We also verify that the MLP non-linearities really matter.
%We replace them with the identify function, i.e.\ making MLPs linear.
%As expected, this clearly underperforms any other choice.

%----------------------------------------------------------------------------------------------
\vspace{-3pt}
\subsection{Compatibility of Optimized Architectures Across Language Datasets}
\label{sec:acrossTasksLang}

\begin{figure}[h!]
    \vspace{-2pt}
    \centering
    \begin{minipage}[t]{0.38\linewidth}
        \setlength{\tabcolsep}{2pt}
        \begin{tabular}{rc}
        \raisebox{60pt}{\makecell[r]{\scriptsize Architectures\\
            \scriptsize optimized for\\
            \scriptsize specific tasks}}&
            \includegraphics[height=1.1\linewidth]{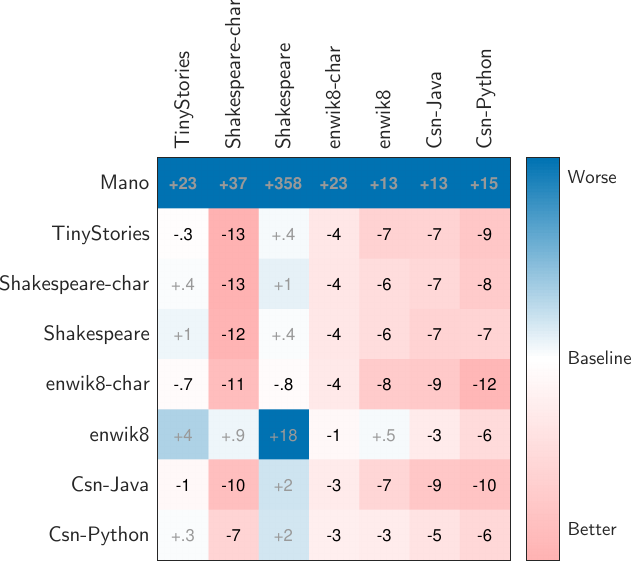}
            \\[2pt]
        ~ & \hspace{13pt}\scriptsize Target tasks
        \end{tabular}
    \end{minipage}%
    \hfill
    \begin{minipage}[t]{0.384\linewidth}
        %\hspace{20pt}
        \vspace{-30pt}
        %\raisebox{15pt}{
        \caption{Compatibility of architectures across code and
        language datasets (relative difference in perplexity with the baseline in~\%, lower is better).
        The differences are much less dramatic than
        with algorithmic tasks (\autoref{fig:acrossTasksAlg}),
        indicating smaller benefit in dataset-specific specialization.}
        \label{fig:acrossTasksLang}
    \end{minipage}
    %\vspace{-8pt}
\end{figure}

Our final results examine the compatibility of the
optimized architectures across language modeling datasets.
We consider our seven %code and natural language
datasets
plus \textsc{Mano}, the most complex of our algorithmic tasks.
%previously studied in the context of pretrained LLM \citep{allenzhu2025canon}
%(whereas our other algorithmic tasks are usually studied on small task-specific models)
We train models for every
task $\mathbb{D}$
using architectures optimized for any other task $\mathbb{D}'$.
%and report in \autoref{fig:acrossTasksLang}
%differences in performance with the baseline architecture.
The results in \autoref{fig:acrossTasksLang}
show that the variations across architectures are very small.
This contrasts with the results on algorithmic tasks (\autoref{fig:acrossTasksAlg}).
These optimized architectures thus encode much less task-specific specialization.
This suggests that the skills required
across code and language modeling datasets
are much more uniform.
We discuss the implications of these results in \autoref{sec:discussion}.
%at least to the extent of what was possibly captured with the proposed method

\vspace{-4pt}

\begin{comment}
\begin{figure}[h!]
    \centering
    \hspace{-55pt}
    \begin{tabular}{rc}
    \raisebox{30pt}{\makecell[r]{\scriptsize Architectures\\
        \scriptsize optimized for\\
        \scriptsize specific tasks}}&
        \includegraphics[height=0.35\linewidth]{figCrossLang/figCrossLang8-2.pdf}
        \\[2pt]
    ~ & \scriptsize Target tasks
    \end{tabular}
    \vspace{-6pt}
    \caption{Compatibility of architectures across code and
    language datasets (relative difference in perplexity with the baseline in \%, lower is better).
    The differences are much less dramatic than
    with algorithmic tasks (\autoref{fig:acrossTasksAlg}),
    indicating smaller benefit in dataset-specific specialization.}
    \label{fig:acrossTasksLang}
\end{figure}
\end{comment}

%\vspace{-9pt}
\clearpage
\myFrame{\textbf{Take-away.}~~For code and natural language modeling,
the optimized architectures
improve %over standard transformers
much less than for algorithmic tasks.
This means that standard transformers are
intrinsically closer to a local optimum in the space of architecture
for these tasks, than for algorithmic skills.}
\vspace{3pt}

\section{Related Work}
\label{sec:relatedWork}
%\vspace{-5pt}

\paragraph{Understanding inductive biases in NNs.}
Much of the prior on understanding neural networks~(NNs) %and transformers
has focused on their simplicity bias, i.e.\ their preference for 
representing functions of low
%sensitivity \citet{hahn2021sensitivity}
Kolmogorov \citep{zhou2023algorithms}
or spectral complexity \citep{bhattamishra2022simplicity}.
The simplicity bias depends primarily
on the choice of activation function \citep{mingard2019neural,teney2024neural},
and its suitability %to various tasks
was questioned \citep{domingos1999role} by evaluating
alternative activation functions in MLPs \citep{teney2024neural}.
We extend this inquiry to transformers and larger settings.
In particular, we introduce a method to optimize non-linearities in both attention and MLP layers,
%of a transformer,
and apply it to tasks relevant to the state of the art (code, natural language, algorithmic reasoning).

%\vspace{-3.5pt}
\textbf{Improving transformers.}
Current LLMs all use very similar architectures,
and %a recent study shows
\citet{mayilvahanan2025llms} show
that small design differences play little role in their performance.
%A recent study suggests that the architecture plays a minor role in the performance of current LLMs \citep{mayilvahanan2025llms}.
%compared to the pretraining data and tokenization method.
Prior work has however studied at length 
the impact of various components of transformers including their nonlinearities
%which is particularly relevant to our work
\citep{jha2025entropy,newhouse2025training}.
%2025 Entropy-Guided Attention for Private LLMs (wo LayerNorms or GeLUs)
%2025 Training Transformers with Enforced Lipschitz Constants
Proposed improvements include alternative attention mechanisms
\citep{katharopoulos2020transformers,saratchandran2024rethinking,schlag2021linear,tamayo2025your,velivckovic2024softmax}
%For example, \citet{velivckovic2024softmax} attribute failure of length generalization to softmax being too ``soft'', making attention scores disperse when length gets longer.
%Softmax is not enough (for sharp out-of-distribution)
and activation functions for MLPs
\citep{hu2025curvature,teney2025we}
and transformers
\citep{mirzadeh2023relu,so2021searching}.
% 2024 ReLU Strikes Back - Exploiting Activation Sparsity in LLMs (ReLUs almost as good as others, with sparsity)
% 2025 Tropical Attention - Neural Algorithmic Reasoning for Combinatorial Algorithms
% 2024 Softmax is not enough for sharp out-of-distribution (lower softmax temp in attention for longer sequences)
% 2021 Searching for efficient transformers for language modeling
% evolutionary search in space of TensorFlow programs; difference in our setting: lower-level (focus on custom activation functions), optimization on held-out data (and for OOD generalization), toy reasoning tasks rather than language modeling; only 1 task, decoder-only language modeling; we explore how the best inductive biases vary across tasks
This %prior work
motivates our work by suggesting that standard transformers
are not a uniquely optimal choice of architecture.

%\vspace{-3.5pt}
\textbf{Architecture search.}
Our method to optimize architectures is reminiscent of
neural architecture search (NAS)
\citep{goyal2019learning,hong2025asnn,liu2018progressive,manessi2018learning,ramachandran2017searching,zoph2016neural}.
The goals and approach are different though.
NAS uses RL or evolutionary algorithms to search through pre-defined design choices.
We directly use gradient descent
to optimize a relatively unrestricted parametrization of the non-linearities
of transformers. Our goal is not to find better models (our designs are often computationally expensive). Instead, our method is a tool to understand the compatibility of the inductive biases required for various tasks.
%See \refApp{app:relatedWork} for an extended literature review.

\begin{comment}
Non-robustness of transformers
  OthelloGPT learned a bag of heuristics
  \citep{nikankin2024arithmetic}
  Arithmetic Without Algorithms: Language Models Solve Math With a Bag of Heuristics, https://arxiv.org/abs/2410.21272
  Evaluating the World Model Implicit in a Generative Model, https://arxiv.org/abs/2406.03689

language models learn natural grammatical structures better than impossible grammars (based on counts, with shuffles/reversals of (parts of) sentences, etc.)
  Mission - Impossible Language Models
\end{comment}

%\vspace{-4pt}
\section{Discussion}
\label{sec:discussion}
%\vspace{-2pt}

We have presented a method to optimize a transformer architecture for specific datasets
and used it to study the compatibility of inductive biases across tasks.
%In many cases, 
We found that standard transformers are often suboptimal, %for many algorithmic tasks,
but minor tweaks (replacing GeLUs and softmax operations)
can substantially improve training speed, generalization, capacity, and stability across random seeds.

%\vspace{-2pt}
Our results show that different tasks benefit from different inductive biases,
aligning with the no-free-lunch theorem \citep{wolpert1996lack}.
Yet, transformers seem uniquely suitable to a vast range of applications \citep{goldblum2023no}:
our results can be seen as probing the limits of this hypothesis.

%\vspace{-1pt}
\textbf{Architecture vs.\ scale.}
Prior work showed that the choice of architecture
%picking the right architecture and inductive biases
can become
less important with scale
\citep{bachmann2023scaling,tay2022scaling}.
But this also means that the current need to build ever-larger models %to improve performance
may be due to %our use of
suboptimal inductive biases.
In this work, we tweaked the non-linearities of 
transformers,
but different inductive biases could be obtained
by completely different means
%e.g.\
such as initializations
\citep{shinnick2025transformers} or optimizers \citep{pascanu2025optimizers}.
%What we show is really the value of different inductive biases, it just happens that tweaking the non-linearities in a transformer is a convenient way to explore the space of possible inductive biases.
%But the same effects (and others) can probably be achieved by other means, e.g. completely different archittectures, or even different means of instilling inductive biases like different initializations, pretraining \citep{shinnick2025transformers}, optimizer, etc.

%\vspace{-1pt}
\textbf{Do we need domain-specific models?}
Our results show a higher compatibility across language and code
than algorithmic tasks.
The latter
are often used to highlight limitations of transformers.
%e.g.\ for length generalization.
If they %algorithmic tasks
really represent desirable capabilities in LLMs,
perhaps new architectures are required to combine capabilities for both language and algorithmic skills.
A future step could be to apply our method to optimize architectures for multiple tasks simultaneously.
It would also be interesting to evaluate whether the improved architectures transfer to other domains such as vision and speech.
%and go beyond the task-specific nature of the improvements demonstrated in this paper.

%\vspace{-1pt}
%\textbf{Other domains.}
%Another extension of this work could examine
%transformers for other domains such as vision and speech,
%and evaluate whether improved architectures transfer across domains.
%and examine the differences between architectures optimized for different domains

% it's curious that we use the ~same architecture (transformers) for such different tasks and datasets (ViTs, LLMs, etc.): is this an optimal choice in any of these cases?

% transformers require much more training data and compute than necessary, e.g. compared to human brains So there is still something missing (in the inductive biases)

% examine better *capacity* to (over)fit tr points \citet{shwartz2024just}

%\vspace{-1pt}
\clearpage
\textbf{Limitations.}
First, our \textbf{search space of architectures} is limited.
Complex forms of attention \citep{hashemi2025tropical}
or interactions like gated linear units (GLU,\,\citealt{shazeer2020glu})
cannot be represented in our formulation.
Further gains are possible with a larger search space,
but the optimization will also be more challenging. %in our experience.
Second, the \textbf{scale of our experiments} is tiny relative to state-of-the-art LLMs.
The effects of different architectures may vanish with more data,
but improving data efficiency is a key objective of this line of work.
%with better the inductive biases is to .
So, effects at a small scale are intrinsically meaningful.
Third, our architectures with optimized non-linearities can seem {computationally costly}.
However, we describe \textbf{efficient implementations} in \autoref{app:implementation}
and find that polynomial approximations of the learned splines make this a non-issue
%Our claims are not centered on the performance of these architecture %per se
%though.
%nor on an incremental improvement of transformers.
%They serve instead to better understand the landscape of possible designs for future AI models.

Code is available at \url{https://github.com/idiap/lm-afs}.

\begin{comment}    
\subsection*{Reproducibility Statement}

Appendix~\ref{app:implementation}
provides a formal description of the proposed method
with the values of all hyperparameters.
%Code is available at \url{http://github.com/anonymized/anonymized}.
\end{comment}



\bibliography{references}
\bibliographystyle{iclr2026_conference}

\clearpage
\appendix

%\section*{Appendix}
\renewcommand{\sectionautorefname}{Appendix}
\renewcommand{\subsectionautorefname}{Appendix}
\renewcommand{\subsubsectionautorefname}{Appendix}

\vspace{-3pt}
\section{Additional Related Work}
\label{app:relatedWork}
\vspace{-1pt}

\textbf{Inductive biases in deep learning}
are due to choices of architecture \citep{goyal2022inductive} and of the learning algorithm (optimizer, objective, regularizers \citealt{kukavcka2017regularization}).
We focus on the former.
The simplicity bias has been studied from both aspects.
Most explanations attribute it to loss functions \citep{pezeshki2021gradient}
and gradient descent
\citep{arora2019implicit,hermann2020shapes,lyu2021gradient,tachet2018learning}.
But work on untrained networks shows %unambiguously
that it can be explained with architectures alone \citep{de2019random,goldblum2023no,mingard2019neural,teney2024neural,valle2018deep}.
\citet{teney2024neural} showed that the choice of activation function can modulate the simplicity bias.
The \textbf{spectral bias} \citep{rahaman2019spectral,kalimeris2019sgd}
or frequency principle \citep{xu2019frequency}
is a related but different
effect related to training dynamics:
NNs approximate low-frequency components of the target function earlier during training with SGD.
%Works on this topic are specific to gradient descent
%\citep{soudry2018implicit,xu2019training}.
%and often to ReLU networks \citep{hong2022activation,huh2021low,zhang2023shallow}.

% New
\vspace{1pt}
\textbf{Simplicity bias in transformers.}
% What Algorithms can Transformers Learn - Study in Length Generalization
%\citet{zhou2023algorithms} explain generalization of transformer models on toy reasoning tasks using a transformer-specific measure of complexity. They propose that the function learned by a transformer corresponds to the shortest program (in a custom programming language) that could generate the training data.
% Simplicity bias in transformers and their ability to learn sparse boolean functions
%\citet{bhattamishra2022simplicity} showed that transformers are more biased for simplicity than LSTMs.
The hypothesis of a simplicity bias in NNs has also been studied specifically in transformers.
\citet{hahn2021sensitivity} shows that common models in NLP are biased to learn low-sensitivity functions.
%2021 Sensitivity as a Complexity Measure for Sequence Classification Tasks (common models biased to learn low sensitivity fcts; sensitivity for non-binary data)
\citet{bhattamishra2022simplicity} shows that transformers are more biased for simplicity than LSTMs.
%2022 Simplicity bias in transformers and their ability to learn sparse boolean functions (transf more biased for simplicity than LSTMs; binary inputs only)
\citet{dziri2023faith} examine large pretrained models and determine that that tend rely on shortcut learning on simple reasoning tasks.
%2023 Faith and Fate - Limits of Transformers on Compositionality (simple tasks, shortcut pattern matching in GPT models)
\citet{zhou2023algorithms} focus on length generalization and show that transformers learn the shortest program in the RASP language that fits the training data --a specific form of the simplicity bias.
%2023 What Algorithms can Transformers Learn - Study in Length Generalization (trsfmers learn the shortest RASP program that fits the tr data)
\citet{rende2024distributional} study BERT-like models and find that they learn simple functions first during the course of training.
%2024 Distributional simplicity bias in the learning dynamics of transformers (about order of learning, simple fct first; BERT-like models)
\citet{zhang2024initialization} find that the scale of initialization can influence a transformer's learning of a generalizing or memorizing solution.
%2024 Initialization is Critical to Whether Transformers Fit Composite Functions by Inference or Memorizing (larger init causes memorization; task not understood)
\citet{vasudeva2024simplicity} further study the bias of transformers for learning low-sensitivity functions using the NTK theory.
%2024 Simplicity Bias of Transformers to Learn Low Sensitivity Functions (theory using NTK and experiments)
\citet{hahn2024sensitive} show that sensitive functions are hard to learn for transformers because they correspond to sharp solutions in their optimization landscape as a side-effect of the simplicity bias.
%2024 Why are Sensitive Functions Hard for Transformers (complex function are sharp minima)

%2025 Transformers Learn Low Sensitivity Functions - Investigations and Implications (trf more sensitive; wrt embeddings, not actual inputs)
%2025 When do Neural Networks Learn World Models (low-order bias helps)

\vspace{1pt}
\textbf{Activation functions}
are key for introducing non-linearities in NNs.
Many options were considered early on, e.g.\ 
sine activations in the Fourier Neural Networks from 1988 \citep{gallant1988there}.
ReLUs are often credited for enabling the rise of deep learning
by avoiding vanishing gradients \citep{maas2013rectifier}.
However they are also essential in inducing the simplicity bias \citep{teney2024neural}
which may be just as important. %for the success of NNs.
%insensitive to network depth and weight magnitudes.
The research community has slowly converged towards smooth handcrafted variants of ReLUs
such as GeLUs \citep{dubey2022activation,hendrycks2016gaussian,ramachandran2017swish}.
%State-of-the-art architectures today use hand-crafted variants such as GeLUs \citep{dubey2022activation,hendrycks2016gaussian,ramachandran2017swish}
%that the research community has converged on empirically.
%by trial and error.
%\citep{dubey2022activation} %Activation functions in deep learning A comprehensive survey and benchmark
%already shows that different AFs are suitable to different types of data and tasks
%we go a step further by allowing to learn very different AFs, e.g. with discontinuities or sharp transitions
Some works proposed \textbf{learning activation functions}
%by parameterizing them
using extra parameters optimized alongside the weights of the network \citep{alexandridis2025adaptive,apicella2019simple,apicella2021survey,bingham2020evolutionary,chelly2024trainable,ducotterd2024improving,jagtap2020adaptive,scardapane2019kafnets,sutfeld2020adaptive}.
See \citet{jagtap2023important} for a comprehensive review.
%, such that they can evolve during training.
%The method in this paper differs in three ways.
The goal is to better fit the training data
with an activation function that can evolve during training.
In contrast, we use meta learning to find an activation function that induces better inductive biases,
such that training with this \emph{fixed} activation provides better generalization.
This requires bi-level optimization, episodic training, and unbiased parametrization
%of the activation function
that allows us to learn activations very different from existing ones.
\textbf{Kolmogorov-Arnold Networks} \citep{liu2024kan}
%is an architecture developed concurrently with this work
parametrize the connections in a NN, which is equivalent to learning different activation functions
across channels and layers.
They use a parametrization as splines similar to ours.
Their benefits in physics-related problems likely result from the alterations to the inductive biases studied in this paper.
Our method differs from \textbf{neural architecture search} \citep{white2023neural}
%(NAS)
in its ability to discover novel activation functions from scratch, rather than selecting
from predefined candidates \citep{sutfeld2020adaptive}
or from a narrow set of parametric functions \citep{alexandridis2025adaptive}.

\vspace{1pt}
\textbf{Length generalization}
refers to the ability of a model to generalize to sequences longer than seen during training,
especially for algorithmic tasks (e.g.\ arithmetic operations on numbers with more digits).
This remains a challenge despite extensive work
%that focuses mostly focus 
on positional encodings,
which only partially address the problem
\citep{anil2022exploring,kazemnejad2023impact,zhou2024transformers}.
%on modified components of transformers
%that (only partially) address the challenge.
This paper shows that other aspects of the architecture can be important.
We use the \textsc{Copy} task as proof of concept
and show that different MLP activation functions can bring a significant improvement
to the existing Alibi encodings
\citep{press2021train}.

%(\autoref{sec:lengthGeneralization})\liang{explain why it is a `even' here? For example, by saying that the failures of length generalization are mostly attributed to the positional encoding but not (never?) activation functions (see reference in comments)}.
% Paper 1: The Impact of Positional Encoding on Length Generalization in Transformers
% Paper 2: Exploring Length Generalization in Large Language Models
% Paper 3: Transformers Can Achieve Length Generalization But Not Robustly

\clearpage

\textbf{Learnability and inductive biases.}
The learnability of any given task is a fundamental question in machine learning.
It is well know that inductive biases are indispensable for generalization to unseen data \citep{mitchell1980need}
and that no learning algorithm is universally useful, as per one of the no-free lunch theorems~\citep{wolpert2002supervised}).
Meanwhile, neural networks have nevertheless proved widely successful. The broad applicability of transformers, in particular, suggests that their inductive bias has a broad relevance to real-world data \citep{goldblum2023no}.
The \textbf{simplicity bias} is a broad and vague characterization of these properties.
Various studies have established however that the simplicity bias is not universally beneficial \citep{domingos1999role,teney2025we,zeng2023transformers}
and even responsible
for failure cases such as 
\textbf{shortcut learning}~\citep{geirhos2020shortcut,puli2023don,teney2021evading}
or the amplification of biases and performance disparities~\citep{bell2023simplicity}.
Even the underlying principle supporting the simplicity bias, known as \textbf{Occam's razor},
has long been debated in the philosophical literature because it lacks a justification from first principles~\citep[Appendix~A]{mingard2023deep}.
A prominent argument for simplicity is rooted in algorithmic information theory~\citep{dingle2018input}
with results stating  essentially that
``\textit{a bias in the distribution of target functions must be towards \underline{low} complexity}''.
However, this only means that simplicity is a good prior on average,
but not necessarily the best choice for any task or dataset.

Studies in linguistics and cognitive science have also examined the question of learnability.
This includes studies on the influence of architectures and data on generalization during \textbf{language acquisition}, both for humans and machines \citep{futrell2023how,milliere2024language,warstadt2020what}.
This explains how syntactic and structural biases arise and how they can be controlled \citep{mueller2022how,papadimitriou2022injecting,yang2024anything}.
%factors driving failures or unexpected limitations \citep{kallini2024mission}
Our paper complements this line of work since it helps clarify the impact of architectures on generalization.
Our approach is quite different though. Our method allows searching through the space of architectures via the optimization of non-linearities.
This matters because current popular designs (e.g.\ transformers) are contingent on external factors (see the \textit{Hardware Lottery}, \citealt{hooker2021hardware}).

\vspace{6pt}
\textbf{Connection with prior work.}
This paper is a follow-up the study by \citet{teney2025we}
that uses trainable non-linearities to study whether the \textit{simplicity bias} of standard neural architectures is always desirable.
%It left many outstanding questions because 
It was however limited to MLPs and toy data, and relied on an expensive optimization method
unsuitable to modern architectures.
In comparison, our main innovations are:
%\begin{itemize}[leftmargin=*, itemsep=2pt, parsep=0pt, topsep=-5pt] % Less space
\begin{itemize}[leftmargin=*, itemsep=3pt, parsep=0pt, topsep=-3pt] % More space
\item a formulation of trainable non-linearities that applies to transformers' MLPs and attention layers;
\item a tractable optimization method replacing the expensive bi-level approach from prior work;
\item the study of mainstream domains (language modeling, algorithmic reasoning);
\item the study of cross-task compatibility, whereas prior work focuses on individual datasets;
\item a demonstration of massive improvements on algorithmic tasks;
%\item a PyTorch implementation that allows swapping standard activation functions for optimized ones with a few lines of code, available at \url{https://github.com/anonymized/anonymized}.
\end{itemize}
%The reason we do not use the optimization method from \citet{teney2025we} is its computational cost. The meta-learning/bi-level approach is impractical with modern architectures and larger models.
%The approximation that we propose is akin to first-order approximations \citep{} commonly used in other meta learning settings \citep{}.
%[5] Reptile: a scalable metalearning algorithm, Nichol et al. 2018
%[6] Model-agnostic meta-learning for fast adaptation of deep networks, Finn et al. 2017

\clearpage

\section{Implementation Details}
\label{app:implementation}

\paragraph{Proposed method.}
We provide a formal description of our method in \autoref{alg:alg}.
\vspace{-7pt}

\newenvironment{algOuter}[1][]{%
  \mdfsetup{%
     frametitle={%
       \tikz[baseline=(current bounding box.east),outer sep=0pt]
        \node[anchor=east,rectangle,fill=black!2]
        {\strut #1};}}%
   \mdfsetup{innertopmargin=-5pt,linecolor=black!50,
             innerleftmargin=1pt,innerrightmargin=1pt,
             leftmargin=0pt,rightmargin=0pt,
             backgroundcolor=black!2,
             linewidth=0.5pt,topline=true,
             frametitleaboveskip=\dimexpr-\ht\strutbox\relax,}
   \begin{mdframed}[]\relax%
   }{\end{mdframed}}

\newenvironment{algInner}[1][]{%
  \mdfsetup{%
     frametitle={%
       \tikz[baseline=(current bounding box.east),outer sep=0pt]
        \node[anchor=east,rectangle,fill=black!2]
        {\strut #1};}}%
   \mdfsetup{innertopmargin=-5pt,linecolor=black!50,
             innerleftmargin=1pt,innerrightmargin=1pt,
             leftmargin=5pt,rightmargin=5pt,
             backgroundcolor=black!2,
             linewidth=0.5pt,topline=true,
             frametitleaboveskip=\dimexpr-\ht\strutbox\relax,}
   \begin{mdframed}[]\relax%
   }{\end{mdframed}}

\setlength{\textfloatsep}{0pt} % Small spacing after algorithm

\begin{algorithm}[th]
    \caption{Proposed method (stage~\rom{1}) to optimize a transformer architecture for a specific task.}
    \label{alg:alg}
    \setlength{\baselineskip}{\baselineskip+1pt}
    \vspace{3pt}
    \textbf{Input}: \\
    \phantom{I}Training data
    $\mathbb{D} = \{\bs_i\}_{i=1}^{n}$ ~as %a set of
    token sequences $\bs\in\bS$. \\
    \phantom{I}Baseline architecture $\mathcal{T}$
    instantiable as next-token prediction model
    \phantom{I}$T_\btheta\!: \bS \!\to\! \bS$
    ~of weights $\btheta$.\\
    \phantom{I}$\mathcal{L}(\cdot,\cdot)$: \,Loss function.%\\    
    %\phantom{I}
    ~~~~
    $\alpha$: \,Fraction of held-out data.%\\
    %\phantom{I}
    ~~~~
    $M$: \,Number of parallel models.\\[3pt]
    \textbf{Method}:\\
    \phantom{I}Define a new architecture
    $\mathcal{\hat{T}}_{\,\btheta_\mathrm{A} \, \btheta_\mathrm{MLP} }$
    by replacing\\
    \phantom{I}~$-$ ~softmaxes with~
    ${\Sigma_j\, K(\bQ_i, \bK_j)\,\bV_j}
    \, \big/ \,
    {\Sigma_j\, K(\bQ_i, \bK_j)},$
    where
    $K(\bQ,\bK) \!=\! \phi_{\btheta_\mathrm{A}}(\bQ)^{\top}\phi_{\btheta_\mathrm{A}}(\bK)$.\\
    \phantom{I}~$-$ ~GeLUs with a linear spline $\phi_{\btheta_\mathrm{MLP}},$\\
    \phantom{I}where architecture hyperparameters $\btheta_\mathrm{A}$ and $\btheta_\mathrm{MLP}$
    specify the value of splines $\phi$ at their keypoints.\\[1.5pt]
    \phantom{I}Instantiate $M$ untrained models of architecture $\mathcal{\hat{T}}$ as $\hat{T}^1_{\btheta_1}$ ... $\hat{T}^M_{\btheta_M}$.\\[1.5pt]
    \phantom{I}Split $\mathbb{D}$ into %disjoints subsets
    $\mathbb{D}_\mathrm{arch}$
    and
    $\mathbb{D}_\mathrm{wts}$
    of sizes 
    $\alpha\,n$
    and
    $(1\!\!-\!\alpha) n$.
    %to optimize weights and architecture. %respectively.
    \\[-11pt]
    \begin{algOuter}[\textbf{while} \textnormal{{not converged}}
    \algComment{SGD training loop}
    ]
        \setlength{\baselineskip}{\baselineskip+1pt}
        \phantom{I}Sample mini-batch $\mathbb{D}^0$ from $\mathbb{D}_\mathrm{arch}$ \, and $\mathbb{D}^1$ ... $\mathbb{D}^M$ from $\mathbb{D}_\mathrm{wts}$
        %\algComment{Different data ordering per model}
        \\[2pt]
        \phantom{I}Eval.\ loss of each individual model on its own data $\mathbb{D}^m$: $L^m_\mathrm{wts}\leftarrow
            \Sigma_{\bs\,\in \mathbb{D}^m}\,
            \mathcal{L}\big(
            \hat{T}^m_{\btheta_m}(\bs), \bs
            \big)$\\
        \phantom{I}Eval.\ combined loss of all models together on $\mathbb{D}^0$:
            \hspace{6pt}
            $L_\mathrm{arch}\leftarrow
            \Sigma_m \,
            \Sigma_{\bs\,\in \mathbb{D}^0}\,
            \mathcal{L}\big(
            \hat{T}^m_{\btheta_m}(\bs), \bs
            \big)$\\[2pt]
        \phantom{I}Update weights of each model: $\forall m, \,\btheta_m \leftarrow \operatorname{SGD}(\btheta_m, \nabla_\btheta L^m_\mathrm{wts})$\\[2.5pt]
        \phantom{I}Update architecture: $
        (\btheta_\mathrm{A}, \btheta_\mathrm{MLP})
        \leftarrow
        \operatorname{SGD}(
        (\btheta_\mathrm{A}, \btheta_\mathrm{MLP})
        , \nabla_{(\btheta_\mathrm{A}, \btheta_\mathrm{MLP})} L_\mathrm{arch})$
    \end{algOuter}
    \vspace{-3pt}
    \phantom{.}\!$(\btheta^\star_\mathrm{A}, \,
    \btheta^\star_\mathrm{MLP})
    ~\leftarrow~
    (\btheta_\mathrm{A}, \,
    \btheta_\mathrm{MLP}).$\\[3pt]
    \textbf{Output}: optimized architecture $\mathcal{\hat{T}}_{\,\btheta^\star_\mathrm{A} \, \btheta^\star_\mathrm{MLP} }$\\%[3pt]
    \phantom{.}\!\!\!\!\algComment{Now $\mathcal{\hat{T}}$ can be used like any other architecture, treating
    $\btheta^\star_\mathrm{A}$ and $\btheta^\star_\mathrm{MLP}$
    as fixed hyperparameters.}
    \vspace{2pt}
\end{algorithm}

\vspace{3pt}
%\clearpage

%----------------------------------------------------------------------------------------------
\paragraph{Baseline transformer architecture.}
Our baseline is a \textsc{GPT-2}-style architecture \citep{radford2019language}.
It uses standard multi-head attention, GeLU activation functions in the MLPs, post-norm layers,
learned absolute positional embeddings, and a width multiplier of $4$ in the MLP hidden layers.
All weights are initialized from Gaussians of standard deviation $0.02$ truncated at $2$ standard deviations.

%----------------------------------------------------------------------------------------------
\phantomsection
\label{sec:splines}
\paragraph{Parametrization of non-linearities as linear splines.}
We want a search space free of priors
such as the smoothness and monotonicity enforced in
similar work on the learning of activation functions (e.g.\ \cite{apicella2019simple,chelly2024trainable}).
We therefore choose to learn a non-linearity as a linear spline
$\phi_\btheta:\mathbb{R}\!\rightarrow\!\mathbb{R}$
with control points defined by $\btheta$.
We define $n_\textrm{c}$ points spread regularly in an interval $[a,b]$,
typically $n_\textrm{c}\!\!=\!\!122$ points in $[-20,+20]$ for a spacing of $1/3$ between points (see hyperparameters in Table~\ref{tab:hyperparameters}).
Then $\phi$ represents piecewise linear segments
interpolating values specified in the learned parameters
$\btheta := [\,\phi_\btheta(a), \ldots \phi_\btheta(b))\,] \in\mathbb{R}^{n_\textrm{c}}$.
The function $\phi$ can represent simple and complex functions,
including smooth curves, periodic functions, sharp transitions, etc.

%----------------------------------------------------------------------------------------------
\begin{comment}
\paragraph{Optimization of architectures.}
Todo.
SWA to make 
fixed-window averaging, typically over the last 20 gradient updates
feasible since our models are small, especially for our experiments on algorithmic toy tasks
\end{comment}

%----------------------------------------------------------------------------------------------
\paragraph{Datasets for algorithmic tasks.}
%For the algorithmic toy tasks, 
For most tasks, we generated data
with code adapted from
\citet{zhong2024algorithmic}:
\url{https://github.com/fjzzq2002/random_transformers}.
While this prior work generates some of the data on-the-fly, we pre-generate all the data to ensure that the training/validation/test splits are strictly disjoint.

For \textsc{Mano}, we re-implemented the data generation based on the description by \citet{allenzhu2025canon}.
Compared to this prior work, we scaled down the task to allow using smaller models.
We generated \num{1e5} training examples, with a number of operations in each sequence in $[1,\!3]$,
a modulus of $7$, and without tokens signaling the number of operations.

For all algorithmic tasks, we use a test set of \num{1e3}
examples, strictly disjoint from the training set.

\paragraph{Datasets for language modeling.}
For datasets tokenized at the character level,
every character or symbol in the data simply corresponds to one token.
For the \textsc{TinyStories}, \textsc{Shakespeare}, and \textsc{enwik8} datasets tokenized at the subword level, we use the 
byte-pair encoding (BPE, \citet{gage1994new}) tokenizer from GPT-2 \citet{radford2019language}.
We consider it a consistent choice suitable to our different datasets
since it was originally trained on very diverse data.
For the \textsc{CodeSearchNet} datasets, we use the tokenizer of the 
CodeGPT model \citep{codegpt}.
For each dataset,
we discard tokens with fewer than 200 training occurrences.
This significantly reduces the vocabulary size and training costs.
This should not undermine the results of our experiments:
if anything, including more rare tokens could reveal larger differences across datasets.

%----------------------------------------------------------------------------------------------
\textbf{Metrics.}
For the algorithmic tasks, we measure performance as the
token-wise accuracy of the ``output'' part of the generated sequences (the same part of the sequences as used to compute the training loss). This allows a finer-grained evaluation of partial success than the sequence-wise accuracy.

For the \textsc{Copy} task, we use the sequence-wise accuracy because the token-wise accuracy
can remain falsely high when a model fails at length generalization.

For the language modeling tasks,
we measure performance using the training perplexity (exponential of cross-entropy loss)
as well as token-wise accuracy on validation data as a more intuitive measure of performance.
For the accuracy, we measure it on the latter half of the context window to ensure that we evaluate predictions with enough conditioning. %to be somewhat deterministic.

For the compatibility across algorithmic tasks (\autoref{fig:acrossTasksAlg}),
we plot the difference in test accuracy with the baseline after a fixed number of steps.
We adapt the number of steps to each task
to capture improvements in generalization and/or training speed depending on the task.
This is because both the baseline and optimized architectures saturate at perfect accuracy
for multiple tasks, hence the \emph{final} accuracy alone is not informative.
\begin{itemize}[leftmargin=*, itemsep=0.5pt, parsep=0pt, topsep=-4pt] % Less space
\item \textsc{memorize}: $150$ steps.
\item \textsc{parentheses}: $300$ steps.
\item \textsc{AddMod}: $300$ steps.
\item \textsc{Haystack}: $400$ steps.
\item \textsc{Add}: $700$ steps.
\item \textsc{AddReversed}: $350$ steps.
\item \textsc{Copy}: $2,\!000$ steps.
\item \textsc{Mano}: $3,\!000$ steps.
\end{itemize}
\vspace{3pt}

%----------------------------------------------------------------------------------------------
%\paragraph{Model and training hyperparameters per task.}
\textbf{Hyperparameters.}
We tuned the hyperparameters in
Table~\ref{tab:hyperparameters}
for a standard transformer on each task,
to make sure that our optimized architectures
are compared against strong baselines.
%the baseline architecture for each task, not for our optimized non-linearities.
%We experimented with re-tuning the learning rate (LR) but did not see any further improvement.
For example, we use the Canon layers proposed by \citet{allenzhu2025canon} for many tasks (sequence-wise 1D convolutions) because they clearly improve the performance of the baseline.

\begin{table}[h!]
    \vspace{-4pt}
    \centering
    \caption{Hyperparameters used for each task.}
    \label{tab:hyperparameters}
    \vspace{-4pt}
    \setlength{\tabcolsep}{2.5pt}
    \renewcommand{\arraystretch}{0.8}
    \begin{tabular}{l | c|c|c|c|c|c| c| c | c |}
    %\begin{tabular}{l | cccccc c c | c |}
    \toprule
    ~&
    \textsc{memorize}&
    \textsc{parentheses}&
    \textsc{AddMod}&
    \textsc{Haystack}&
    \textsc{Add}&
    \textsc{AddReversed}&

    \textsc{Copy}&
    \textsc{Mano}&

    Language datasets\\
    \midrule
    \!Num.\ layers & \multicolumn{8}{c|}{2} & 4\\
    \midrule
    \!Num.\ att.\ heads & 2 & 2 & 2 & 2 & 2 & 2 & 8 & 4 & 4\\
    \midrule
    \!Width & 32 & 32 & 32 & 128 & 128 & 128 & 128 & 128 & 512\\
    \midrule
    \!Tied embeddings & \multicolumn{8}{c|}{No} & Yes\\
    \midrule
    \!Canon layers & \multicolumn{6}{c|}{No} & \multicolumn{3}{c|}{Yes} \\
    \midrule
    \!Num.\ tr.\ steps & 500 & 500 & 1,000 & 1,000 & 1,000 & 1,000 & 2,000 & 5,000 & 3,000 \\
    \midrule
    \!Peak LR & .005 & .001 & .02 & .001 & .001 & .001 & .004 & .001 & .001\\
    \midrule
    \!Batch size & \multicolumn{8}{c|}{512} & 64\\
    \midrule
    \!Optimizer & \multicolumn{9}{c|}{Adam}\\
    \midrule
    \!Adam $(\beta_1,\!\beta_2)$ &
    \multicolumn{6}{c|}{(0.9, 0.999)} &
    \multicolumn{2}{c|}{(0.92, 0.98)} &
    (0.9, 0.999) \\
    \midrule
    \!LR schedule & \multicolumn{9}{c|}{5\% linear warm-up, 50\% cosine cool-down (not necessary for algorithmic tasks; used on all tasks for consistency)}\\
    \midrule
    \!Weight decay & \multicolumn{9}{c|}{0 (better on all tasks than using any weight decay)}\\
    \midrule
    \!Dropout rate & \multicolumn{9}{c|}{0}\\
    \midrule
    \!Parallel models $M$ & \multicolumn{8}{c|}{8} & 3\\
    \midrule
    \!Spline range $[a,\!b]$ & \multicolumn{9}{c|}{$[-20,20]$}\\
    \midrule
    \!Spline spacing $n_\textrm{c}$ &
    \multicolumn{6}{c|}{1/9} &
    \multicolumn{3}{c|}{1/3} \\
    \bottomrule
    \end{tabular}
    \vspace{-5pt}
\end{table}

\clearpage

%----------------------------------------------------------------------------------------------
\section{Additional Results on Algorithmic Tasks}

\begin{figure}[h!]
    \centering
    \setlength{\tabcolsep}{7pt}
    \renewcommand{\arraystretch}{1.2}
    \begin{tabular}{r ccc}
    \textbf{Attention:} & \smax & \smax & Ours\\
    \textbf{MLP:}       & \gelu & Ours  & Ours\\[3pt]
    
    \raisebox{27pt}{\textsc{Memorize}}&
\includegraphics[height=0.135\linewidth]{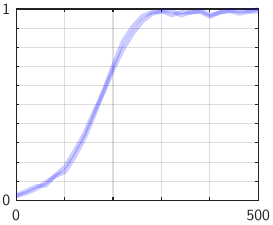}&
\includegraphics[height=0.135\linewidth]{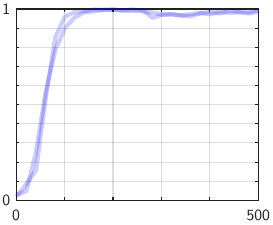}&
\includegraphics[height=0.135\linewidth]{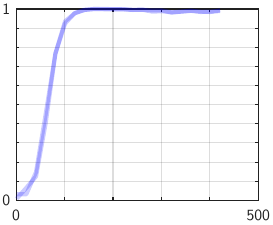}\\

    \raisebox{27pt}{\textsc{Parentheses}}&
\includegraphics[height=0.135\linewidth]{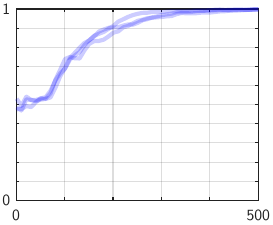}&
\includegraphics[height=0.135\linewidth]{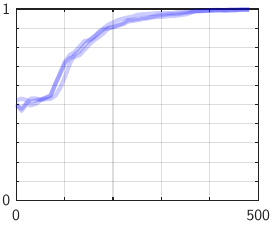}&
\includegraphics[height=0.135\linewidth]{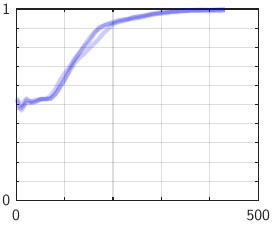}\\

    \raisebox{27pt}{\textsc{AddMod}}&
\includegraphics[height=0.135\linewidth]{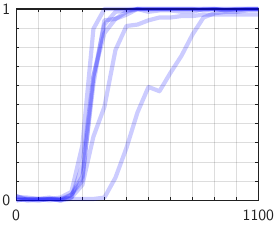}&
\includegraphics[height=0.135\linewidth]{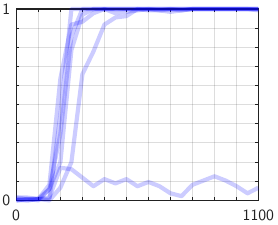}&
\includegraphics[height=0.135\linewidth]{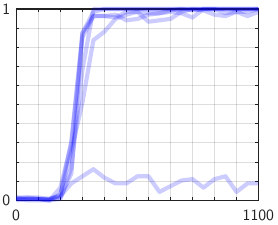}\\

    \raisebox{27pt}{\textsc{Haystack}}&
\includegraphics[height=0.135\linewidth]{figCurvesAlgPdf/curve07-haystack-0-1.pdf}&
\includegraphics[height=0.135\linewidth]{figCurvesAlgPdf/curve07-haystack-1-1.pdf}&
\includegraphics[height=0.135\linewidth]{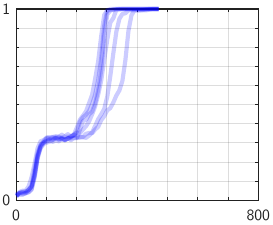}\\

    \raisebox{27pt}{\textsc{Add}}&
\includegraphics[height=0.135\linewidth]{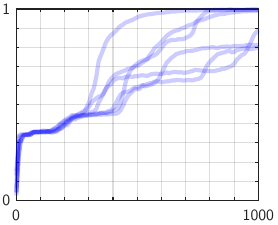}&
\includegraphics[height=0.135\linewidth]{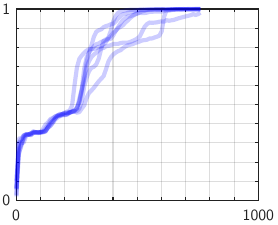}&
\includegraphics[height=0.135\linewidth]{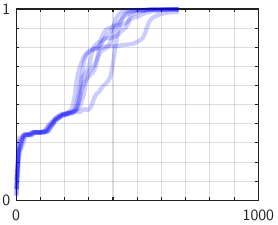}\\

    \raisebox{27pt}{\textsc{AddReversed}}&
\includegraphics[height=0.135\linewidth]{figCurvesAlgPdf/curve10-decimalAddition-0-1.pdf}&
\includegraphics[height=0.135\linewidth]{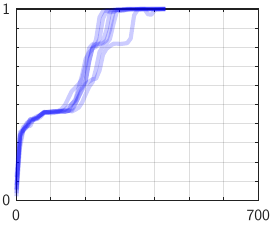}&
\includegraphics[height=0.135\linewidth]{figCurvesAlgPdf/curve10-decimalAddition-3-1.pdf}\\

    \raisebox{27pt}{\textsc{Copy}}&
\includegraphics[height=0.135\linewidth]{figCurvesAlgPdf/curve13-copy-0-1.pdf}&
\includegraphics[height=0.135\linewidth]{figCurvesAlgPdf/curve13-copy-1-1.pdf}&
\includegraphics[height=0.135\linewidth]{figCurvesAlgPdf/curve13-copy-1-1.pdf}\\

    \raisebox{27pt}{\textsc{Mano}}&
\includegraphics[height=0.135\linewidth]{figCurvesAlgPdf/curve30-polMano-0-1.pdf}&
\includegraphics[height=0.135\linewidth]{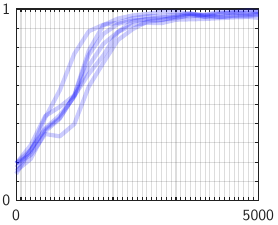}&
\includegraphics[height=0.135\linewidth]{figCurvesAlgPdf/curve30-polMano-3-1.pdf}
    \end{tabular}
    \vspace{-2pt}
    \caption{Training curves (test accuracy vs.\ training steps, one curve per seed) of models trained on algorithmic tasks
    with a baseline transformer (first column) or optimized architectures (second and third columns).}
    \label{fig:trCurvesAlg}
    \vspace{6pt}
\end{figure}

\phantomsection
\label{app:trCurvesAlg}
\textbf{Training curves.}
\autoref{fig:trCurvesAlg} shows that
the optimized architectures (2nd and 3rd columns) always converge significantly faster than 
a baseline transformer (1st column) and show less variance across seeds. There is little difference between the 2nd and 3rd columns, which means that most of the benefits come from optimizing the non-linearity within the MLP layers rather than the attention.
\vspace{6pt}

\clearpage

%----------------------------------------------------------------------------------------------
\phantomsection
\label{app:algCapacity}
\textbf{Compatibility of architectures across algorithmic tasks.}
We present below the full results following the format
of \autoref{fig:acrossTasksAlg}.
We show the effect when optimizing the non-linearities in MLP or attention layers, or both.
Optimizing the non-linearities in the attention proves to be really challenging,
and the best results are usually obtained by optimizing only the MLPs.

\begin{figure}[h!]
    \centering
    \hspace{-49pt}
    \setlength{\tabcolsep}{4pt}
    \begin{tabular}{rccc}
    \raisebox{48pt}{\makecell[r]{\scriptsize Architectures\\
        \scriptsize optimized for\\
        \scriptsize specific tasks}}&
    \includegraphics[trim=0pt 0pt 58pt 0pt,clip,height=0.25\linewidth]{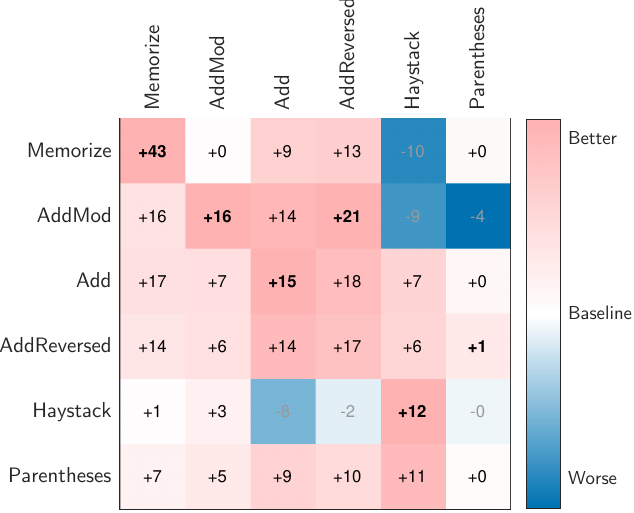}&
    \includegraphics[trim=55pt 0pt 58pt 0pt,clip,height=0.25\linewidth]{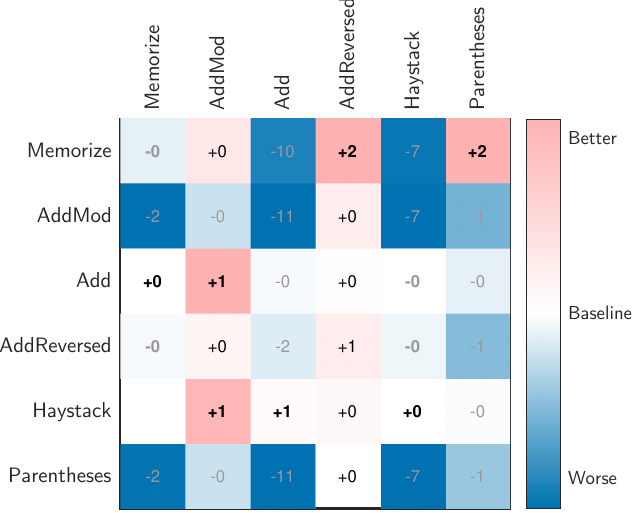}&
    \includegraphics[trim=55pt 0pt 0pt 0pt,clip,height=0.25\linewidth]{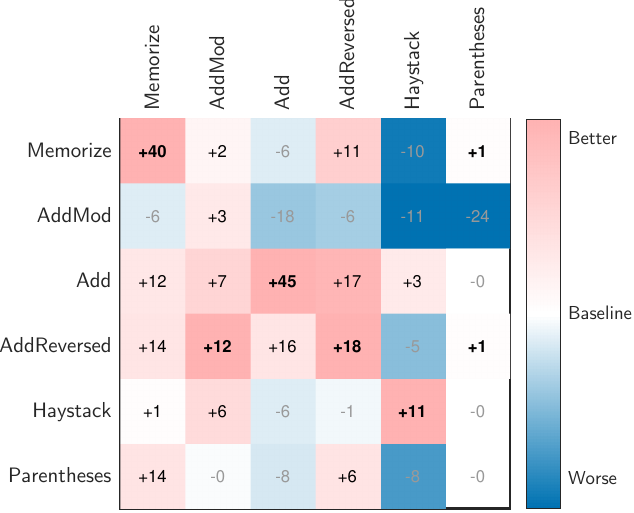}\\[1pt]
    ~ & \hspace{16pt}\scriptsize Target tasks & \scriptsize Target tasks & \scriptsize Target tasks\hspace{20pt}\phantom{.}\\[4pt]
    ~ & \hspace{16pt}\scriptsize \textbf{Optimized MLPs} & \scriptsize \textbf{Optimized att.} & \scriptsize \textbf{Optimized MLP\,\&\,att.}\hspace{20pt}
    \end{tabular}
    \vspace{-4pt}
    \caption{Compatibility of architectures across
    algorithmic tasks (difference in test accuracy with the baseline after a fixed number of steps).}
    \label{fig:acrossTasksAlgFull}
    \vspace{3pt}
\end{figure}

%----------------------------------------------------------------------------------------------
\begin{comment}
\phantomsection
\label{app:algCapacity}
\textbf{Improved capacity.}
See Table \ref{tab:algCapacity} below.

\begin{table}[ht!]
    \centering
    \caption{Improved capacity on algorithmic tasks. Todo.}
    \label{tab:algCapacity}
    \vspace{-3pt}
    \setlength{\tabcolsep}{2.2pt}
    \renewcommand{\arraystretch}{1.0}
    \begin{tabular}{l ccc | cc | ccccccccc}
    \toprule
    Todo.\\
    \bottomrule
    \end{tabular}
    \vspace{6pt}
\end{table}
\end{comment}

%----------------------------------------------------------------------------------------------
\begin{comment}
\phantomsection
\label{app:mdl}
\textbf{Analysis from the MDL perspective.}
Todo.
\end{comment}

\clearpage

%----------------------------------------------------------------------------------------------
\section{Additional Results on Language Modeling}

\begin{figure}[h!]
    \centering
    \setlength{\tabcolsep}{1pt}
    \begin{tabular}{l ccc ccc ccc}
    ~ &
    \tiny 1 layer, dim.\,128 &
    \tiny 1 layer, dim.\,256 &
    \tiny 1 layer, dim.\,512 &
    \tiny 2 layers, dim.\,128 &
    \tiny 2 layers, dim.\,256 &
    \tiny 2 layers, dim.\,512 &
    \tiny 4 layers, dim.\,128 &
    \tiny 4 layers, dim.\,256 \\[4pt]
    \raisebox{19pt}{\tiny \textbf{Original}} &
    \includegraphics[trim=0pt 137pt 0pt 0pt,clip,width=0.11\linewidth]{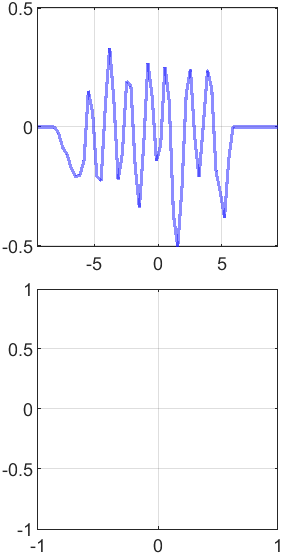} &
    \includegraphics[trim=0pt 137pt 0pt 0pt,clip,width=0.11\linewidth]{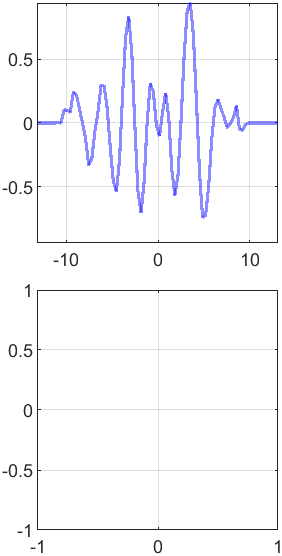} &
    \includegraphics[trim=0pt 137pt 0pt 0pt,clip,width=0.11\linewidth]{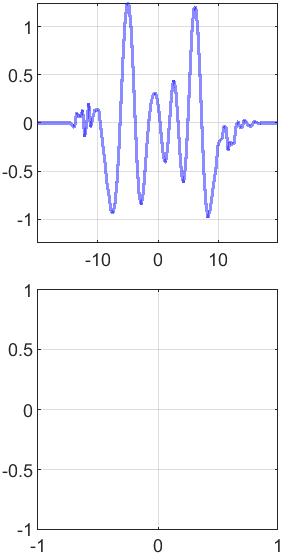} &
    \includegraphics[trim=0pt 137pt 0pt 0pt,clip,width=0.11\linewidth]{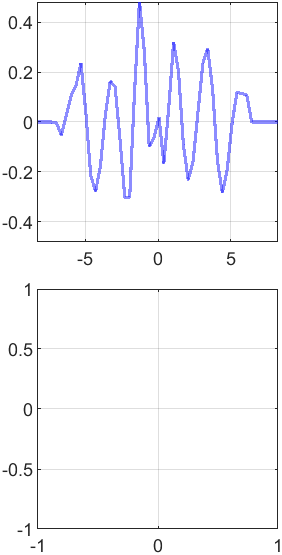} &
    \includegraphics[trim=0pt 137pt 0pt 0pt,clip,width=0.11\linewidth]{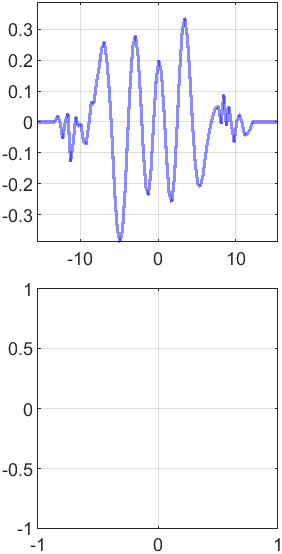} &
    \includegraphics[trim=0pt 137pt 0pt 0pt,clip,width=0.11\linewidth]{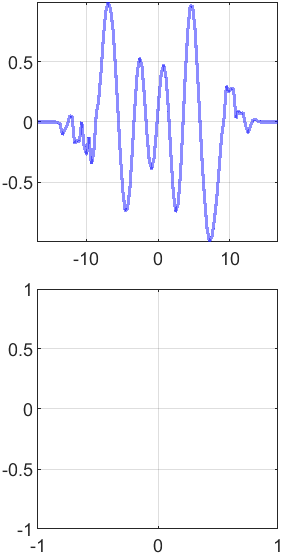} &
    \includegraphics[trim=0pt 137pt 0pt 0pt,clip,width=0.11\linewidth]{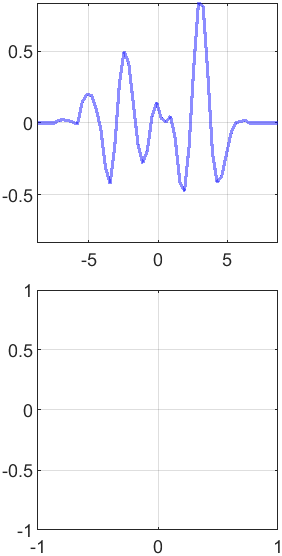} &
    \includegraphics[trim=0pt 137pt 0pt 0pt,clip,width=0.11\linewidth]{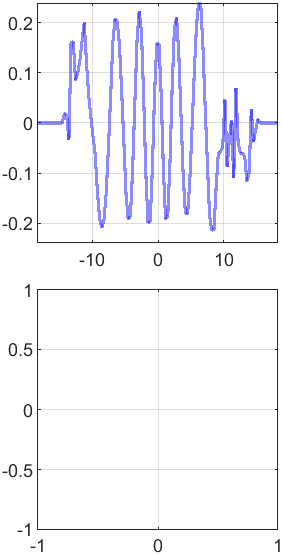} \\
    \tiny Test acc. (\%) &
    59.0 & 62.0 & 63.6 & 60.8 & 64.3 & 65.6 & 62.5 & 65.5 \\[10pt]
    \raisebox{19pt}{\tiny \textbf{Modified}} &
    \includegraphics[trim=0pt 137pt 0pt 0pt,clip,width=0.11\linewidth]{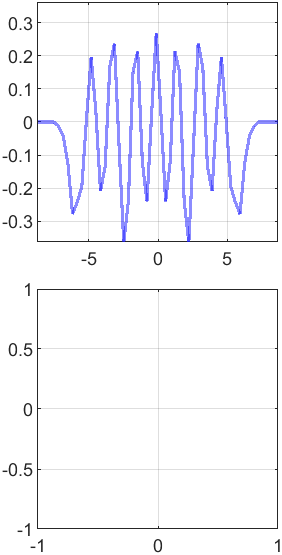} &
    \includegraphics[trim=0pt 137pt 0pt 0pt,clip,width=0.11\linewidth]{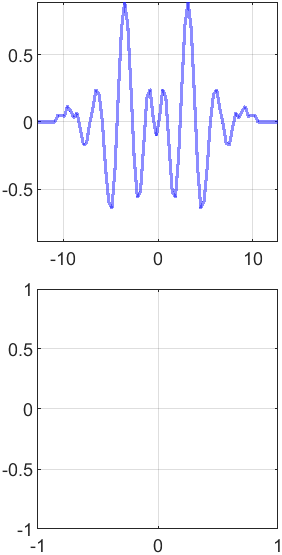} &
    \includegraphics[trim=0pt 137pt 0pt 0pt,clip,width=0.11\linewidth]{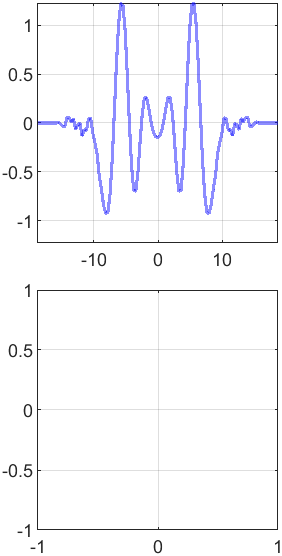} &
    \includegraphics[trim=0pt 137pt 0pt 0pt,clip,width=0.11\linewidth]{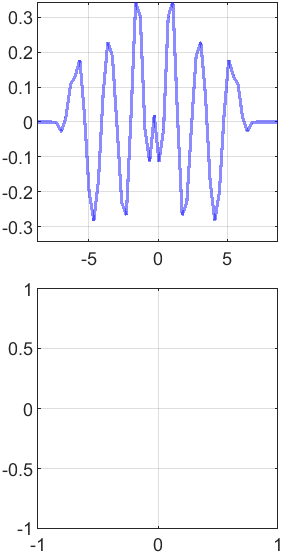} &
    \includegraphics[trim=0pt 137pt 0pt 0pt,clip,width=0.11\linewidth]{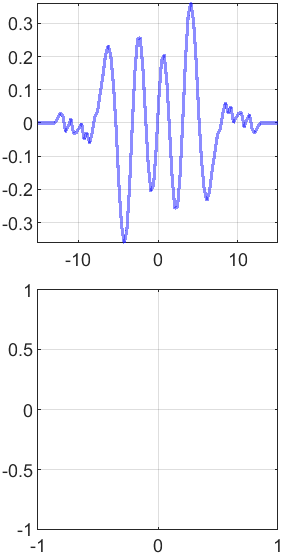} &
    \includegraphics[trim=0pt 137pt 0pt 0pt,clip,width=0.11\linewidth]{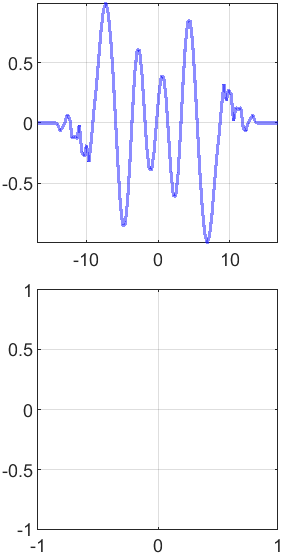} &
    \includegraphics[trim=0pt 137pt 0pt 0pt,clip,width=0.11\linewidth]{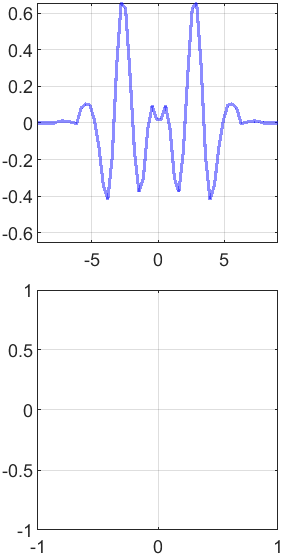} &
    \includegraphics[trim=0pt 137pt 0pt 0pt,clip,width=0.11\linewidth]{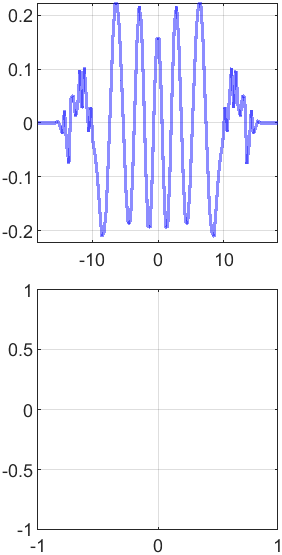} \\
    \tiny Test acc. (\%) &
    58.8 & 61.8 & 63.4 & 60.9 & 64.0 & 65.8 & 62.1 & 65.4\\[1pt]
    ~ & \textcolor{gray}{(-$0.2$)\,} & \textcolor{gray}{(-$0.2$)\,} & \textcolor{gray}{(-$0.2$)\,} & \textcolor{gray}{(+$0.1$)\,} & \textcolor{gray}{(-$0.3$)\,} & \textcolor{gray}{(+$0.2$)\,} & \textcolor{gray}{(-$0.4$)\,} & \textcolor{gray}{(-$0.1$)\,}
    %~ & $-0.2$ & $-0.2$ & $-0.2$ & $+0.1$ & $-0.3$ & $+0.2$ & $-0.2$ & $-0.4$ & $-0.1$
    \end{tabular}
    \vspace{-2pt}
    \caption{
    MLP non-linearities optimized for \textsc{TinyStories}
    and versions modified to enforce symmetry.
    Almost all of these perform worse than the original ones, whose fine details therefore matter.}
    \label{fig:cleanAfs}
    \vspace{8pt}
\end{figure}

%----------------------------------------------------------------------------------------------
\phantomsection
\label{app:cleanAfs}
\textbf{Manipulating optimized non-linearities.}
In these experiments, we slightly modify the optimized MLP non-linearities
to understand the importance of their fine details.
Since %most optimized non-linearities
they often look like sinusoidal wavelets,
perhaps an even more regular version of them could perform better.
We automate a ``cleaning'' process of the optimized non-linearities as follows.
We take the optimized spline, reverse it along the \textsc{x} and/or \textsc{y} axis (yielding three different versions), then align it with the original one by maximizing their cross-correlation.
We then keep the average of the two.
Among the three versions, we retain the one with the highest cross-correlation (i.e.\ similarity)
with the original spline. The result is symmetric or anti-symmetric with fewer irregularities than the original one.
We visualize this effect in \autoref{fig:cleanAfs} on MLP non-linearities optimized
for \textsc{TinyStories} and various model sizes. We train models with these, but in almost every case, they perform worse than the original ones.
This shows that fine details in the original optimized non-linearities matter.

\begin{comment}
Therefore we apply several transformations
that enforce symmetry and/or better periodicity,
which makes them look arguably nicer (see \autoref{fig:cleanAfs}).
We then train models with these modified components.
In all cases, they perform worse than the original one.
This indicates that the irregularities of fine details
of the original optimized non-linearities
do matter.
\end{comment}

% C:\Dropbox\Research\2024-12 Transformers\Results final\results-tinyStories-200-64\clean1 (worse)
    
%----------------------------------------------------------------------------------------------
\phantomsection
\label{app:multimodel}
\paragraph{Multi-model training.}
We compare in \autoref{tab:tinyStoriesNModels} architectures
for \textsc{TinyStories}
obtained with the proposed method
and $M\!\!=\!\!1$ or $M\!\!=\!\!6$ models in parallel.
%(corresponding to \textit{Attention:\,smax / MLP:\,Ours} in \autoref{tab:tinyStories}).
The latter are slightly better, and the optimized non-linearities look slightly more regular.
% C:\Dropbox\Research\2024-12 Transformers\Results final\results-tinyStories-200-64\nModels6

% 2 ly, 256
% 1.59 & 1.57
% 63.78 & 64.30 

% 4 ly, 256
% 1.53 & 1.52
% 65.32 & 65.45

\vspace{-4pt}
\begin{table}[ht!]
    \centering
    \caption{Models for \textsc{TinyStories} with architectures optimized
    %using the proposed method
    with $M\!\!=\!\!1$ or $6$ parallel models.}
    \label{tab:tinyStoriesNModels}
    %\vspace{-1pt}
    \setlength{\tabcolsep}{6pt}
    \renewcommand{\arraystretch}{1.0}
    \raisebox{8pt}{\makecell[l]{\scriptsize (Models with\\[-3pt]\scriptsize \textcolor{blue}{~$2$ layers},\\[-3pt]\scriptsize ~width $256$)}}\hspace{12pt}
    \begin{tabular}{l cc cc}
    \toprule
    \hspace{-3pt}\textbf{Attention}&
      \rot{\smax} &
      \rot{\smax} & \rot{\smax} & \rot{\smax}\\
    \hspace{-3pt}\textbf{MLP}&
      \rot{Linear} &
      \rot{\gelu} & \rot{\!\!\!Ours,\,$M$=$1$\!\!\!} & \rot{\!\!\!Ours,\,$M$=$6$\!\!\!}\\
    \midrule
    \hspace{-3pt}Tr.\ perplexity &
    1.78 & 1.58 & 1.59 & \textbf{1.57}\\
    \hspace{-3pt}Val.\ acc.\ ($\%$) &
    59.9 & 63.7 & 63.8 & \textbf{64.3} \\
    \bottomrule
    \end{tabular}%
    \raisebox{-22pt}{
    \hspace{10pt}
    \begin{picture}(0,0)\put(19.5,48){\scriptsize $M\!\!=\!\!1$}\end{picture}%
    \includegraphics[trim=0pt 145.5pt 0pt 0pt,clip,width=0.128\linewidth]{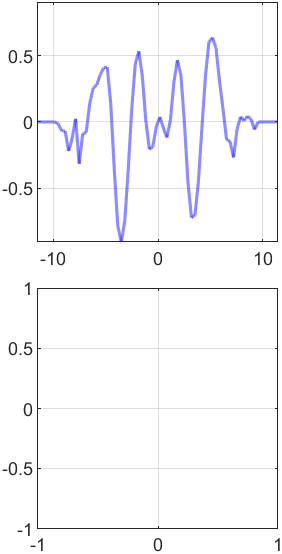}
    \hspace{10pt}
    \begin{picture}(0,0)\put(19.5,48){\scriptsize $M\!\!=\!\!6$}\end{picture}%
    \includegraphics[trim=0pt 145.5pt 0pt 0pt,clip,width=0.128\linewidth]{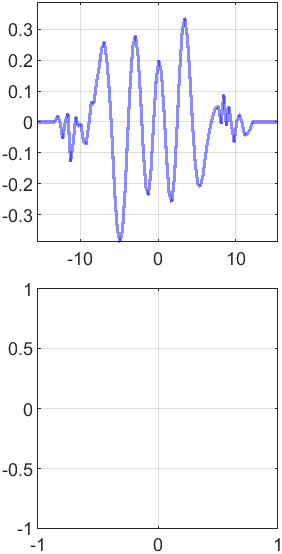}
    }\\\vspace{10pt}
    \raisebox{8pt}{\makecell[l]{\scriptsize (Models with\\[-3pt]\scriptsize \textcolor{blue}{~$4$ layers},\\[-3pt]\scriptsize ~width $256$)}}\hspace{12pt}
    \begin{tabular}{l cc cc}
    \toprule
    \hspace{-3pt}\textbf{Attention}&
      \rot{\smax} &
      \rot{\smax} & \rot{\smax} & \rot{\smax}\\
    \hspace{-3pt}\textbf{MLP}&
      \rot{Linear} &
      \rot{\gelu} & \rot{\!\!\!Ours,\,$N$=$1$\!\!\!} & \rot{\!\!\!Ours,\,$N$=$6$\!\!\!}\\
    \midrule
    \hspace{-3pt}Tr.\ perplexity &
    1.73 & 1.53 & 1.53 & \textbf{1.52}\\
    \hspace{-3pt}Val.\ acc.\ ($\%$) &
    60.8 & 65.1 & 65.3 & \textbf{65.4} \\
    \bottomrule
    \end{tabular}%
    \raisebox{-25pt}{
    \hspace{10pt}
    \includegraphics[trim=0pt 137pt 0pt 0pt,clip,width=0.128\linewidth]{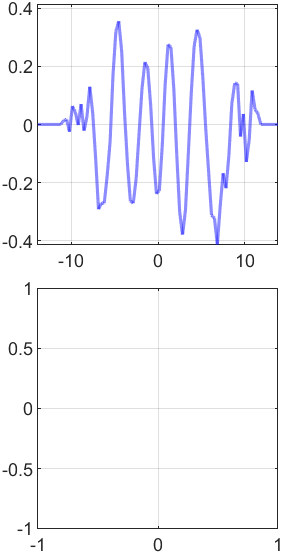}
    \hspace{10pt}
    \includegraphics[trim=0pt 137pt 0pt 0pt,clip,width=0.128\linewidth]{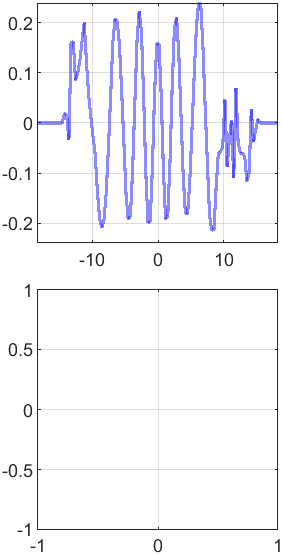}
    }
\end{table}
\vspace{8pt}

%----------------------------------------------------------------------------------------------
\phantomsection
\label{app:alternatives}
\textbf{Existing methods.}
Below are references for 
the attention and MLP designs
evaluated in \autoref{tab:tinyStories}.
%\vspace{-3pt}
%Alternative attention mechanisms:
\begin{itemize}[leftmargin=*, itemsep=3pt, parsep=0pt, topsep=-3pt] % Less space
\item \textbf{Adaptive softmax}: \citet{velivckovic2024softmax}.
\item \textbf{NormSoftmax}: \citet{jiang2023normsoftmax}.
\item \textbf{Polynomial attention P1}: $(\bQ^\top\!\bK) / \sqrt{\mathrm{seqLength}}$:
\citet{saratchandran2024rethinking}.
\item \textbf{Polynomial attention P3}: $(\bQ^\top\!\bK)^3 / \sqrt{\mathrm{seqLength}}$:
\citet{saratchandran2024rethinking}.
%\end{itemize}
%
%\vspace{4pt}
%Alternative MLP designs:
%\begin{itemize}[leftmargin=*, itemsep=3pt, parsep=0pt, topsep=-3pt] % Less space
\item \textbf{GLU}: \citet{shazeer2020glu}.
\item \textbf{ReLU$^2$}: \citet{so2021primer}.
\item \textbf{Sinc}: \citet{saratchandran2024sampling}.
\item \textbf{Gaussian}: \citet{saragadam2023wire}.
\end{itemize}
\vspace{12pt}
\clearpage

%----------------------------------------------------------------------------------------------
\phantomsection
\label{app:shakespeare}
\textbf{Full results on Shakespeare.}
We present below results on the \textsc{Shakespeare} dataset
for various model sizes, in the same format as \autoref{fig:tinyStories}.
The best configuration is to optimize the MLP non-linearities while keeping the original softmax attention (second panels from the left).
%(for character- and subword-level tokenization)

\vspace{6pt}

\begin{figure}[h!]
    \centering
    \setlength{\tabcolsep}{5pt}
    \renewcommand{\arraystretch}{1.0}
    \begin{tabular}{r cccc c}
    ~&
    \includegraphics[trim=0pt 0pt 42pt 0pt,clip,height=0.15\linewidth]{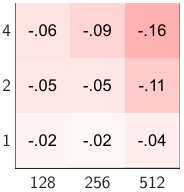}&
    \includegraphics[trim=0pt 0pt 42pt 0pt,clip,height=0.15\linewidth]{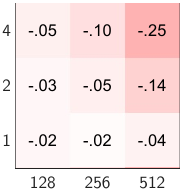}&
    \includegraphics[trim=0pt 0pt 42pt 0pt,clip,height=0.15\linewidth]{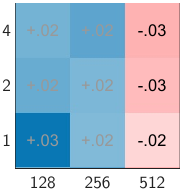}&
    \includegraphics[trim=0pt 0pt 42pt 0pt,clip,height=0.15\linewidth]{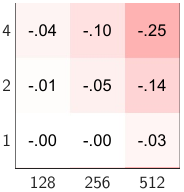}&
    \includegraphics[trim=90pt 0pt 0pt 0pt,clip,height=0.15\linewidth]{figShakespeare/figHeatmap2-shakespeare-interpInterp-2.pdf}\\
    %\midrule
    \textbf{Attention:} & \smax         & \smax & Ours  & Ours \\
    \textbf{MLP:}       & GeLU\,+\,Ours & Ours  & \gelu & Ours
    \end{tabular}
    \vspace{-2pt}
    \caption{Absolute improvements in training perplexity on character-level \textsc{Shakespeare} for models of different sizes (number of layers $\times$ width).}
    \label{fig:shakespeareChar}
\end{figure}

\vspace{10pt}

\begin{figure}[h!]
    \centering
    \setlength{\tabcolsep}{6pt}
    \renewcommand{\arraystretch}{1.0}
    \begin{tabular}{r cccc c}
    ~&
    \includegraphics[trim=0pt 0pt 42pt 0pt,clip,height=0.15\linewidth]{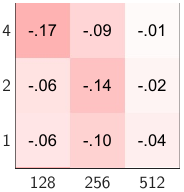}&
    \includegraphics[trim=0pt 0pt 42pt 0pt,clip,height=0.15\linewidth]{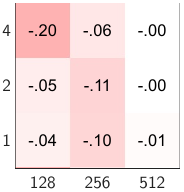}&
    \includegraphics[trim=0pt 0pt 42pt 0pt,clip,height=0.15\linewidth]{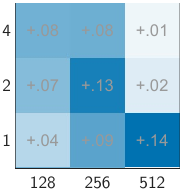}&
    \includegraphics[trim=0pt 0pt 42pt 0pt,clip,height=0.15\linewidth]{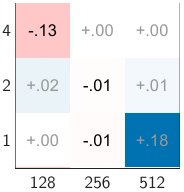}&
    \includegraphics[trim=90pt 0pt 0pt 0pt,clip,height=0.15\linewidth]{figShakespeare/figHeatmap2-shakespeare-interpInterp-2.pdf}\\
    %\midrule
    \textbf{Attention:} & \smax         & \smax & Ours  & Ours \\
    \textbf{MLP:}       & GeLU\,+\,Ours & Ours  & \gelu & Ours
    \end{tabular}
    \vspace{-2pt}
    \caption{Same as \autoref{fig:shakespeareChar} with subword-level tokenization.}
    \label{fig:shakespeareSubword}
    \vspace{3pt}
\end{figure}

\begin{figure}[h!]
    \centering
    \setlength{\tabcolsep}{1.9pt}
    \renewcommand{\arraystretch}{1.0}
    \begin{tabular}{rc rc rc rc}
    \hspace{12pt} & \textsc{TinyStories} & \hspace{18pt} & \textsc{TinyStories} &
    \hspace{18pt} & \textsc{Shakespeare-char} & \hspace{18pt} & \textsc{Shakespeare-char} \\
    \raisebox{6pt}{\rotatebox{90}{\scriptsize Test accuracy (\%)}}~&
    \includegraphics[height=0.15\linewidth]{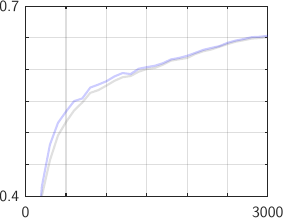}&
    \raisebox{5.5pt}{\rotatebox{90}{\scriptsize Training\ loss ($\log$)}}~&
    \includegraphics[height=0.15\linewidth]{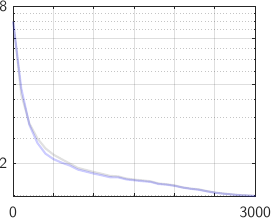}&
    \raisebox{6pt}{\rotatebox{90}{\scriptsize Test accuracy (\%)}}~&
    \includegraphics[height=0.15\linewidth]{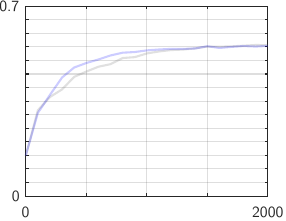}&
    \raisebox{5.5pt}{\rotatebox{90}{\scriptsize Training\ loss ($\log$)}}~&
    \includegraphics[height=0.15\linewidth]{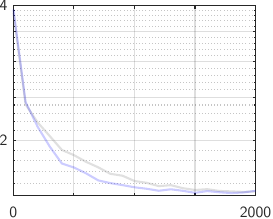}
    \end{tabular}
    \vspace{-2pt}
    \caption{Training curves on language datasets with
    baseline (\lnnn{AAAAAA})
    and
    optimized (\lnnn{9F9FFF})
    architectures.}
    \label{fig:trCurvesLang}
    \vspace{3pt}
\end{figure}

%----------------------------------------------------------------------------------------------
\phantomsection
\label{app:trCuvesLang}
\textbf{Training curves on language datasets.}
\autoref{fig:trCurvesLang} shows that
the optimized architectures
(\lnnn{9F9FFF})
show a larger improvement over a baseline transformer
early during training, which then diminishes.

%----------------------------------------------------------------------------------------------
\clearpage
\section{Results with Larger Language Models}
\label{app:largerModels}
On the suggestions of reviewers, we performed additional experiments %at a larger scale.
to evaluate the improvements from the optimized non-linearities at various scales.
%to verify that the improvements from the optimized non-linearities do not vanish at a larger scale.
%with larger models or more training data.
We repeat experiments on language modeling as in Section~\ref{sec:experimentsLang} with the following differences. %compared to the rest of the paper. %Section~\ref{}.
\begin{itemize}[leftmargin=*, itemsep=3pt, parsep=0pt, topsep=-3pt] % Less space
\item We use the \textbf{\textsc{FineWeb} dataset}~\citep{penedo2024fineweb}, a popular high-quality dataset of cleaned and deduplicated English text from CommonCrawl.
\item We implement our method on top of a very strong baseline, the \textbf{NanoGPT Speedrun}~\citep{modded_nanogpt_2024}.
This is a competitive repository where contributors
specifically push the implementation and data efficiency of the model on the \textsc{FineWeb} dataset.
We specifically build on top of {record \#16}, which includes rotary embeddings, QK normalization, the Muon optimizer, sliding-window attention, mixed-precision training, etc.
The code was designed for 8 H100 GPUs
but we adapted it to enable experiments with a single Nvidia RTX\,4090 laptop GPU.
Our results are therefore not directly comparable with the official Speedrun competition.
%See our code for details:
%\url{https://github.com/anonymized/anonymized}.
\item We first run stage~\rom{1} of our method to optimize the MLP non-linearities of a small model, since this stage is computationally more expensive (2 layers, width 256, 4 attention heads).
We then re-use the optimized non-linearity to run stage~~\rom{2} (i.e.\ standard training) 
with models of \textbf{various sizes from 2 to 12 layers}.
This setup therefore evaluates how the optimized non-linearities transfer across models of different depths.
\item We train similar models (with 2 to 12 layers) with a ReLU, which is the best baseline for this codebase.
We always use a standard attention with a softmax since we found in Section~\ref{sec:experimentsLang} that it was difficult to improve upon.
\end{itemize}
\vspace{8pt}

\textbf{Results}.
The results in Table~\ref{tab:fineWeb} 
show that our optimized non-linearities
perform similarly or better than the baselines.
There is little improvement at the smallest scale (probably because the model is very weak overall) but we get a consistent improvement at all other scales, surpassing both the ReLU and GeLU baselines in most cases.

Regarding the computational cost of the optimized non-linearities,
our implementation (Listing~\ref{lst:spline}) is as fast or faster than a ReLU
in very small models.
In larger models however, they become much more expensive.
We propose in Appendix~\ref{app:efficientApprox} a polynomial approximation.
Table~\ref{tab:fineWeb} shows that this approximation performs
about as well as the original spline
and
about as fast as a ReLU.
%\vspace{8pt}

\begin{table}[ht!]
    %\vspace{-3pt}
    \centering
    \caption{Evaluation of models of various depths trained on \textsc{FineWeb} (average over 3 seeds).}
    \label{tab:fineWeb}
    \vspace{-3pt}
    \setlength{\tabcolsep}{5pt}
    \renewcommand{\arraystretch}{1.03}
    \begin{tabular}{rc}
    Validation loss (\textsc{FineWeb})~~~~~~&
    \begin{tabular}{l ccccc}
    \toprule
    \textbf{Number of layers} & 2 & 4 & 8 & 10 & 12\\
    \color{darkGray}Number of parameters (M) & \color{darkGray}91 & \color{darkGray}105 & \color{darkGray}133 & \color{darkGray}148 & \color{darkGray}162\\
    \midrule
    Linear & ~~4.21 & 4.05 & 3.94 & 3.91 & 3.89 \\
    GeLU & ~~~\textbf{4.00}~~~ & ~~~3.87~~~ & ~~~\textbf{3.72}~~~ & ~~~3.75~~~ & ~~~3.72~~~ \\
    ReLU & ~~~4.01~~~ & ~~~3.87~~~ & ~~~3.78~~~ & ~~~3.75~~~ & ~~~3.72~~~ \\
    \midrule
    Ours: linear spline &
    ~~\textbf{4.00} & \textbf{3.82} & \textbf{3.72} & \textbf{3.70} & 3.68 \\
    Ours: polynomial approx.\ ($n\!=\!18$) &
    ~~4.02 & \textbf{3.82} & \textbf{3.73} & 3.71 & \textbf{3.69} \\
    \bottomrule
    \end{tabular}
    \vspace{8pt}
    \\
    Training time (sec)~~~~~~&
    \begin{tabular}{l ccccc}
    \toprule
    \textbf{Number of layers} & 2 & 4 & 8 & 10 & 12\\
    %\color{darkGray}Number of parameters (M) & \color{darkGray}91 & \color{darkGray}105 & \color{darkGray}133 & \color{darkGray}148 & \color{darkGray}162\\
    \midrule
    Linear & ~~\textbf{1,440} & \textbf{1,920} & \textbf{2,940} & \textbf{3,540} & \textbf{19,680} \\
    GeLU & ~~\textbf{1,440} & \textbf{1,920} & 3,090 & 20,580 & 34,020 \\
    ReLU & ~~{1,500} ~&~ 1,980 ~&~ {3,120} ~&~ {13,080} ~&~     
{28,020} ~\\
    \midrule
    Ours: linear spline &
    ~~{1,500} & 2,070 & 8,520 & 26,700 & 81,720\\
    Ours: polynomial approx.\ ($n\!=\!18$) &
    ~~\textbf{1,440} & {2,040} & {3,180} & {14,070} & {29,100}\\
    \bottomrule
    \end{tabular}
    \end{tabular}\\
    \vspace{8pt}
    \centering
    \includegraphics[width=0.58\textwidth]{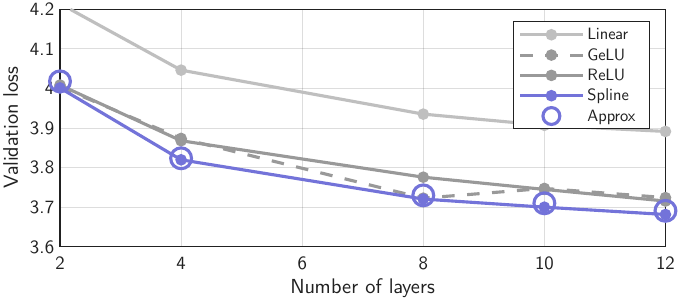}
    \vspace{-15pt}
\end{table}

%----------------------------------------------------------------------------------------------
\clearpage
\section{Efficient Implementation of Splines}
\label{app:efficientApprox}
%\vspace{-3pt}

\textbf{Exact implementation.}
Our non-linearities are parametrized as linear splines.
We first provide
an exact efficient implementation (Listing~\ref{lst:spline})
that we find to be as fast as standard activations such as GeLUs for small models.
However, depending on the architecture and GPU used,
this function can quickly get bandwidth-constrained and become significantly slower.
Therefore we propose a faster approximation with polynomials
to be used when the spline has already been optimized and is used as a frozen non-linearity (i.e.\ for standard training, as in stage~\rom{2} of our experiments).

\begin{listing}[h!]
    \vspace{-5pt}
    \begin{mdframed}[style=frameStyleListing,userdefinedwidth=
    .93\linewidth,innerrightmargin=3pt,innerleftmargin=-7pt,leftmargin=12pt,rightmargin=0pt]
\begin{MintedVerbatim}[commandchars=\\\{\}]
    \PYG{c+c1}{\PYGZsh{} Evaluate at points x a function defined as the linear interpolation of knots}
    \PYG{c+c1}{\PYGZsh{} of coordinates \PYGZsq{}knotPos\PYGZsq{} and values \PYGZsq{}knotVals\PYGZsq{}.}
    \PYG{n+nd}{@torch}\PYG{o}{.}\PYG{n}{compile}\PYG{p}{(}\PYG{n}{dynamic}\PYG{o}{=}\PYG{k+kc}{False}\PYG{p}{)}
    \PYG{k}{def}\PYG{+w}{ }\PYG{n+nf}{eval\PYGZus{}spline}\PYG{p}{(}\PYG{n}{x}\PYG{p}{,} \PYG{n}{knotPos}\PYG{p}{,} \PYG{n}{knotVals}\PYG{p}{):}  \PYG{c+c1}{\PYGZsh{} x: bfloat16, knotPos/knotVals: float32}
        \PYG{n}{idx} \PYG{o}{=} \PYG{n}{torch}\PYG{o}{.}\PYG{n}{bucketize}\PYG{p}{(}\PYG{n}{x}\PYG{p}{,} \PYG{n}{knotPos}\PYG{p}{)} \PYG{o}{\PYGZhy{}} \PYG{l+m+mi}{1}  \PYG{c+c1}{\PYGZsh{} Find the interval each x falls into}
        \PYG{n}{idx} \PYG{o}{=} \PYG{n}{idx}\PYG{o}{.}\PYG{n}{clamp}\PYG{p}{(}\PYG{l+m+mi}{0}\PYG{p}{,} \PYG{n+nb}{len}\PYG{p}{(}\PYG{n}{knotPos}\PYG{p}{)} \PYG{o}{\PYGZhy{}} \PYG{l+m+mi}{2}\PYG{p}{)}
        \PYG{n}{stepSize} \PYG{o}{=} \PYG{n}{knotPos}\PYG{p}{[}\PYG{l+m+mi}{1}\PYG{p}{]} \PYG{o}{\PYGZhy{}} \PYG{n}{knotPos}\PYG{p}{[}\PYG{l+m+mi}{0}\PYG{p}{]}  \PYG{c+c1}{\PYGZsh{} Interval between adjacent knots}
        \PYG{n}{x0} \PYG{o}{=} \PYG{n}{knotPos}\PYG{p}{[}\PYG{l+m+mi}{0}\PYG{p}{]} \PYG{o}{+} \PYG{n}{idx} \PYG{o}{*} \PYG{n}{stepSize}  \PYG{c+c1}{\PYGZsh{} Closest knot \PYGZlt{}= x}
        \PYG{n}{frac} \PYG{o}{=} \PYG{p}{(}\PYG{n}{x} \PYG{o}{\PYGZhy{}} \PYG{n}{x0}\PYG{p}{)} \PYG{o}{/} \PYG{n}{stepSize}  \PYG{c+c1}{\PYGZsh{} Fractional position of x within the interval}
        \PYG{n}{frac} \PYG{o}{=} \PYG{n}{frac}\PYG{o}{.}\PYG{n}{clamp}\PYG{p}{(}\PYG{l+m+mf}{0.0}\PYG{p}{,} \PYG{l+m+mf}{1.0}\PYG{p}{)}  \PYG{c+c1}{\PYGZsh{} Ensure constant extrapolation beyond the knots}
        \PYG{n}{y0} \PYG{o}{=} \PYG{n}{knotVals}\PYG{p}{[}\PYG{n}{idx}\PYG{p}{]}      \PYG{c+c1}{\PYGZsh{} Value of preceding knot (\PYGZlt{}= x)}
        \PYG{n}{y1} \PYG{o}{=} \PYG{n}{knotVals}\PYG{p}{[}\PYG{n}{idx} \PYG{o}{+} \PYG{l+m+mi}{1}\PYG{p}{]}  \PYG{c+c1}{\PYGZsh{} Value of following knot (\PYGZgt{} x)}
        \PYG{n}{out} \PYG{o}{=} \PYG{n}{y0} \PYG{o}{+} \PYG{n}{frac} \PYG{o}{*} \PYG{p}{(}\PYG{n}{y1} \PYG{o}{\PYGZhy{}} \PYG{n}{y0}\PYG{p}{)}  \PYG{c+c1}{\PYGZsh{} Linear interpolation}
        \PYG{k}{return} \PYG{n}{out}\PYG{o}{.}\PYG{n}{to}\PYG{p}{(}\PYG{n}{x}\PYG{o}{.}\PYG{n}{dtype}\PYG{p}{)}  \PYG{c+c1}{\PYGZsh{} For mixed precision; x can be bfloat16, NOT knotPos}
\end{MintedVerbatim}
\begin{comment}
    \begin{minted}[fontsize=\scriptsize]{python}
    # Evaluate at points x a function defined as the linear interpolation of knots
    # of coordinates 'knotPos' and values 'knotVals'.
    @torch.compile(dynamic=False)
    def eval_spline(x, knotPos, knotVals):  # x: bfloat16, knotPos/knotVals: float32
        idx = torch.bucketize(x, knotPos) - 1  # Find the interval each x falls into
        idx = idx.clamp(0, len(knotPos) - 2)
        stepSize = knotPos[1] - knotPos[0]  # Interval between adjacent knots
        x0 = knotPos[0] + idx * stepSize  # Closest knot <= x
        frac = (x - x0) / stepSize  # Fractional position of x within the interval
        frac = frac.clamp(0.0, 1.0)  # Ensure constant extrapolation beyond the knots
        y0 = knotVals[idx]      # Value of preceding knot (<= x)
        y1 = knotVals[idx + 1]  # Value of following knot (> x)
        out = y0 + frac * (y1 - y0)  # Linear interpolation
        return out.to(x.dtype)  # For mixed precision; x can be bfloat16, NOT knotPos
    \end{minted}
\end{comment}
    \end{mdframed}
    \vspace{-5pt}
    \caption{Exact evaluation of a linear spline, used for stages~\rom{1} and~\rom{2} of most of our experiments.\label{lst:spline}}
\end{listing}

\textbf{Approximation with polynomials.}
The splines learned in our experiments with language models are quite smooth
(unlike with algorithmic tasks in Section~\ref{sec:experimentsAlg}).
It is therefore reasonable to approximate them with polynomials,
which are much simpler and faster to evaluate.
Concretely, given a linear spline optimized in stage~\rom{1} of our method,
we determine an approximation through a least-squares fit of a polynomial of chosen degree $n$
on the spline values at its knots, on its support that has non-zero values.
We choose a high degree ($n\!=\!18$ typically)
to ensure high fidelity with the original spline and to avoid ringing artifacts near the support boundaries.
Beyond the boundaries, the polynomial is clamped to 0.
For efficiency, we evaluate the polynomial with Horner's method, and implement it in a compiled function using \texttt{TorchsSscript} (see Listing~\ref{lst:poly}).

\begin{listing}[h!]
    \vspace{-5pt}
    \begin{mdframed}[style=frameStyleListing,userdefinedwidth=
    .98\linewidth,innerrightmargin=5pt,innerleftmargin=-8pt,leftmargin=-8pt,rightmargin=0pt]
    %\centering
\begin{MintedVerbatim}[commandchars=\\\{\}]
    \PYG{n+nd}{@torch}\PYG{o}{.}\PYG{n}{jit}\PYG{o}{.}\PYG{n}{script}
    \PYG{k}{def}\PYG{+w}{ }\PYG{n+nf}{eval\PYGZus{}polynomial}\PYG{p}{(}\PYG{n}{x}\PYG{p}{:} \PYG{n}{torch}\PYG{o}{.}\PYG{n}{Tensor}\PYG{p}{)} \PYG{o}{\PYGZhy{}\PYGZgt{}} \PYG{n}{torch}\PYG{o}{.}\PYG{n}{Tensor}\PYG{p}{:}
        \PYG{n}{x} \PYG{o}{=} \PYG{n}{x}\PYG{o}{.}\PYG{n}{clamp}\PYG{p}{(}\PYG{o}{\PYGZhy{}}\PYG{l+m+mf}{79.52}\PYG{p}{,} \PYG{l+m+mf}{71.65}\PYG{p}{)} \PYG{c+c1}{\PYGZsh{} Clamp for constant extrapolation}
        \PYG{n}{x} \PYG{o}{=} \PYG{n}{x} \PYG{o}{/} \PYG{l+m+mf}{79.52} \PYG{c+c1}{\PYGZsh{} Normalize to get values within [\PYGZhy{}1,1] for numerical stability}
        \PYG{k}{return} \PYG{p}{((((((((((((((((((}\PYG{l+m+mf}{29327.20}\PYG{p}{)}\PYG{o}{*}\PYG{n}{x} \PYG{o}{+} \PYG{l+m+mf}{18324.92}\PYG{p}{)}\PYG{o}{*}\PYG{n}{x} \PYG{o}{\PYGZhy{}} \PYG{l+m+mf}{41591.43}\PYG{p}{)}\PYG{o}{*}\PYG{n}{x} \PYG{o}{\PYGZhy{}} \PYG{l+m+mf}{12376.90}\PYG{p}{)}\PYG{o}{*}\PYG{n}{x} \PYG{o}{\PYGZhy{}}
        \PYG{l+m+mf}{14822.88}\PYG{p}{)}\PYG{o}{*}\PYG{n}{x} \PYG{o}{\PYGZhy{}} \PYG{l+m+mf}{29015.27}\PYG{p}{)}\PYG{o}{*}\PYG{n}{x} \PYG{o}{+} \PYG{l+m+mf}{33452.63}\PYG{p}{)}\PYG{o}{*}\PYG{n}{x} \PYG{o}{+} \PYG{l+m+mf}{10354.57}\PYG{p}{)}\PYG{o}{*}\PYG{n}{x} \PYG{o}{+} \PYG{l+m+mf}{21105.54}\PYG{p}{)}\PYG{o}{*}\PYG{n}{x} \PYG{o}{+} \PYG{l+m+mf}{45592.25}\PYG{p}{)}\PYG{o}{*}\PYG{n}{x} \PYG{o}{\PYGZhy{}}
        \PYG{l+m+mf}{47565.33}\PYG{p}{)}\PYG{o}{*}\PYG{n}{x} \PYG{o}{\PYGZhy{}} \PYG{l+m+mf}{47925.56}\PYG{p}{)}\PYG{o}{*}\PYG{n}{x} \PYG{o}{+} \PYG{l+m+mf}{26296.37}\PYG{p}{)}\PYG{o}{*}\PYG{n}{x} \PYG{o}{+} \PYG{l+m+mf}{18216.14}\PYG{p}{)}\PYG{o}{*}\PYG{n}{x} \PYG{o}{\PYGZhy{}} \PYG{l+m+mf}{6145.61}\PYG{p}{)}\PYG{o}{*}\PYG{n}{x} \PYG{o}{\PYGZhy{}} \PYG{l+m+mf}{2660.53}\PYG{p}{)}\PYG{o}{*}\PYG{n}{x} \PYG{o}{+}
        \PYG{l+m+mf}{522.13}\PYG{p}{)}\PYG{o}{*}\PYG{n}{x} \PYG{o}{+} \PYG{l+m+mf}{66.86}\PYG{p}{)}\PYG{o}{*}\PYG{n}{x} \PYG{o}{\PYGZhy{}} \PYG{l+m+mf}{0.63} \PYG{c+c1}{\PYGZsh{} Evaluate polynomial with Horner\PYGZsq{}s method}
\end{MintedVerbatim}
\begin{comment}
    \begin{minted}[fontsize=\scriptsize]{python}
    @torch.jit.script
    def eval_polynomial(x: torch.Tensor) -> torch.Tensor:
        x = x.clamp(-79.52, 71.65) # Clamp for constant extrapolation
        x = x / 79.52 # Normalize to get values within [-1,1] for numerical stability
        return ((((((((((((((((((29327.20)*x + 18324.92)*x - 41591.43)*x - 12376.90)*x - 
        14822.88)*x - 29015.27)*x + 33452.63)*x + 10354.57)*x + 21105.54)*x + 45592.25)*x - 
        47565.33)*x - 47925.56)*x + 26296.37)*x + 18216.14)*x - 6145.61)*x - 2660.53)*x + 
        522.13)*x + 66.86)*x - 0.63 # Evaluate polynomial with Horner's method
    \end{minted}
\end{comment}
    \end{mdframed}
    \vspace{-5pt}
    \caption{Example of a polynomial approximation, for the best spline from Table~\ref{tab:afApprox}. It uses Horner's method with hard-coded coefficients and is compiled with \texttt{TorchScript} for efficiency.
    Note that this function was learned for the \textit{NanoGPT speedrun} codebase
    which has unusually large internal activations, and is unlikely to directly work well with other architectures.
    \label{lst:poly}}%
    %from Figure~\ref{TODO} (degree 18).}
\end{listing}

% Exact code: (truncated to 2 decimals above for readability)
%import torch
%@torch.jit.script
%def evalPoly(x: torch.Tensor) -> torch.Tensor:
%    x = x.clamp(-79.52755737304688, 79.52755737304688) / 79.52755737304688
%    return ((((((((((((((((((-10986.16796875) * x + -21883.8671875) * x + -10175.6396484375) * x + 54885.953125) * x + 153546.3125) * x + -8314.5732421875) * x + -327620.1875) * x + -105438.71875) * x + 331085.71875) * x + 144341.734375) * x + -185560.46875) * x + -85531.40625) * x + 58941.50390625) * x + 25047.203125) * x + -9917.7919921875) * x + -3184.41943359375) * x + 688.5792236328125) * x + 78.09406280517578) * x + -1.850327730178833

\textbf{Importance of \emph{high degree} polynomials.}
We tried reducing the maximum degree of the polynomials.
This creates smoother functions that look appealing, but they perform systematically worse
than high-degree polynomials or than the original spline.
This shows the importance of fine details in the optimized splines.
We also tried to suppress noise and artifacts near the support boundaries,
by analytically enforcing null derivatives (up to 4th derivatives) of the polynomial at the boundaries.
The functions are again visually appealing but they do not necessarily work better when training models with them.
The data-driven optimization is clearly superior to our hand-crafted tweaks.
One possible improvement that we have not implemented is an approximation with Chebyshev polynomials.
These are known to provide better approximations of functions with finite supports, with less artifacts and better numerical stability.

\textbf{Do we need splines at all?}
We tried to do away with splines entirely and directly optimize coefficients of a polynomial
in stage~\rom{1} of our method. This completely fails however.
Even though splines and polynomials can represent similar sets of functions,
the different parametrization apply different inductive biases on the learned non-linearities.
As discussed in Section~\ref{sec:methods}, splines are particularly effective
because they correspond to the most uniform prior on the space of functions.
%impose the fewest constraints
%offer the weakest inductive bias

\clearpage
\textbf{Evaluation of polynomial approximations.}
We train small language models on \textsc{FineWeb}
with a different non-linearity for the MLP layers.
We keep all hyperparameters identical
%across models (not specifically tuned for our non-linearities)
and similar to Section~\ref{app:largerModels}.
Here, we use 6 layers, a width of 256, 4 attention heads, $\sim$20M parameters, and $\sim$80M training tokens.
%\textbf{Results.}
The results
in Table~\ref{tab:afApprox}
%the ReLU and linear baselines
%with our splines and approximations thereof.
show that our spline performs best and slightly better than a ReLU.
As expected, \textbf{the polynomial approximations are increasingly effective
as we increase the degree}. The approximation then becomes very close to the exact spline.
Low-degree polynomials yield smoother functions that are visually appealing but do not work as well.
This shows that the parametrization as a spline is important to capture subtle important details.

% Right-aligned, vertically centered m-column
\newcolumntype{R}[1]{>{\raggedleft\arraybackslash}m{#1}}
\newcolumntype{L}[1]{>{}m{#1}}

\begin{table}[h!]
    \vspace{-3pt}
    %\centering
    \caption{Models trained on \textsc{FineWeb} with various MLP non-linearities.
    Our optimized spline works best.
    Approximations with high-degree polynomials are effective as they faithfully approximate the spline.}
    \label{tab:afApprox}
    \renewcommand{\arraystretch}{1.0}
    \setlength{\tabcolsep}{5pt}
    \vspace{-3pt}
    \begin{tabular}{r L{7cm}}
        %\toprule
        %\textbf{MLP non-linearity} 
        & \hspace{9em}\textbf{Validation loss (\textsc{FineWeb})}\\%[3pt]
        %\midrule
        \begin{tabular}{R{2.1cm} R{1cm}}
        %\begin{tabular}{c c}
            Linear & \includegraphics[width=0.0955\textwidth]{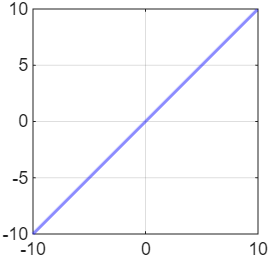}\\
            ReLU & \includegraphics[width=0.0955\textwidth]{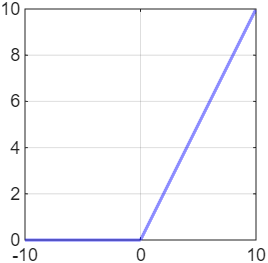}\\
            \textbf{Spline (exact)} & \includegraphics[width=0.0955\textwidth]{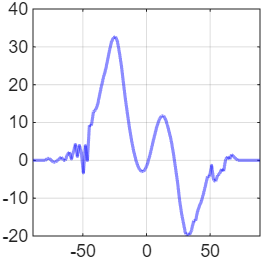}\\
            Approx.\ ($n\!=\!3$) & \includegraphics[width=0.0955\textwidth]{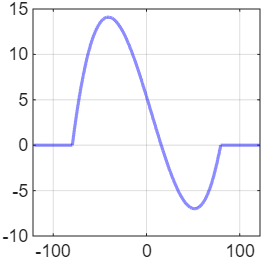}\\
            Approx.\ ($n\!=\!4$) & \includegraphics[width=0.0955\textwidth]{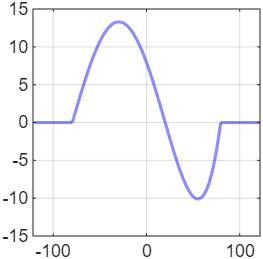}\\
            Approx.\ ($n\!=\!5$) & \includegraphics[width=0.0955\textwidth]{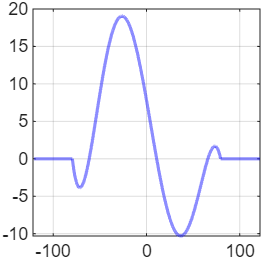}\\
            %Approx.\ ($n\!=\!6$) & \includegraphics[width=0.0955\textwidth]{figAfApprox/af-f1l6d256h4tb1-n-50-0-spline-s1-re000010-approx6.1-step000000.png}\\
            Approx.\ ($n\!=\!10$) & \includegraphics[width=0.0955\textwidth]{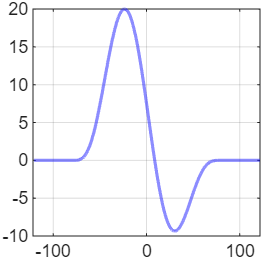}\\
            Approx.\ ($n\!=\!12$) & \includegraphics[width=0.0955\textwidth]{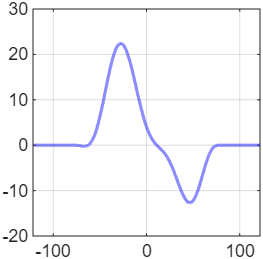}\\
            Approx.\ ($n\!=\!14$) & \includegraphics[width=0.0955\textwidth]{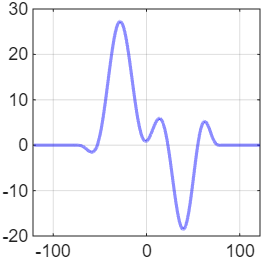}\\
            Approx.\ ($n\!=\!15$) & \includegraphics[width=0.0955\textwidth]{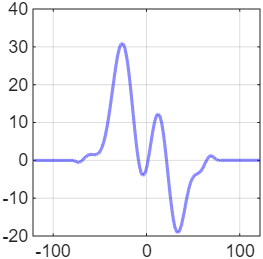}\\
            \textbf{Approx.\ ($n\!=\!17$)} & \includegraphics[width=0.0955\textwidth]{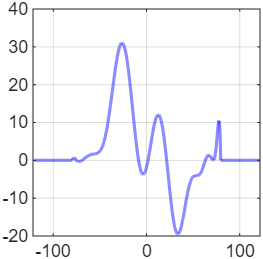}\\
            Approx.\ ($n\!=\!18$) & \includegraphics[width=0.0955\textwidth]{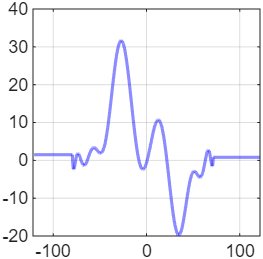}\\[18.5pt]
        \end{tabular}
        &
          \includegraphics[height=17.31cm]{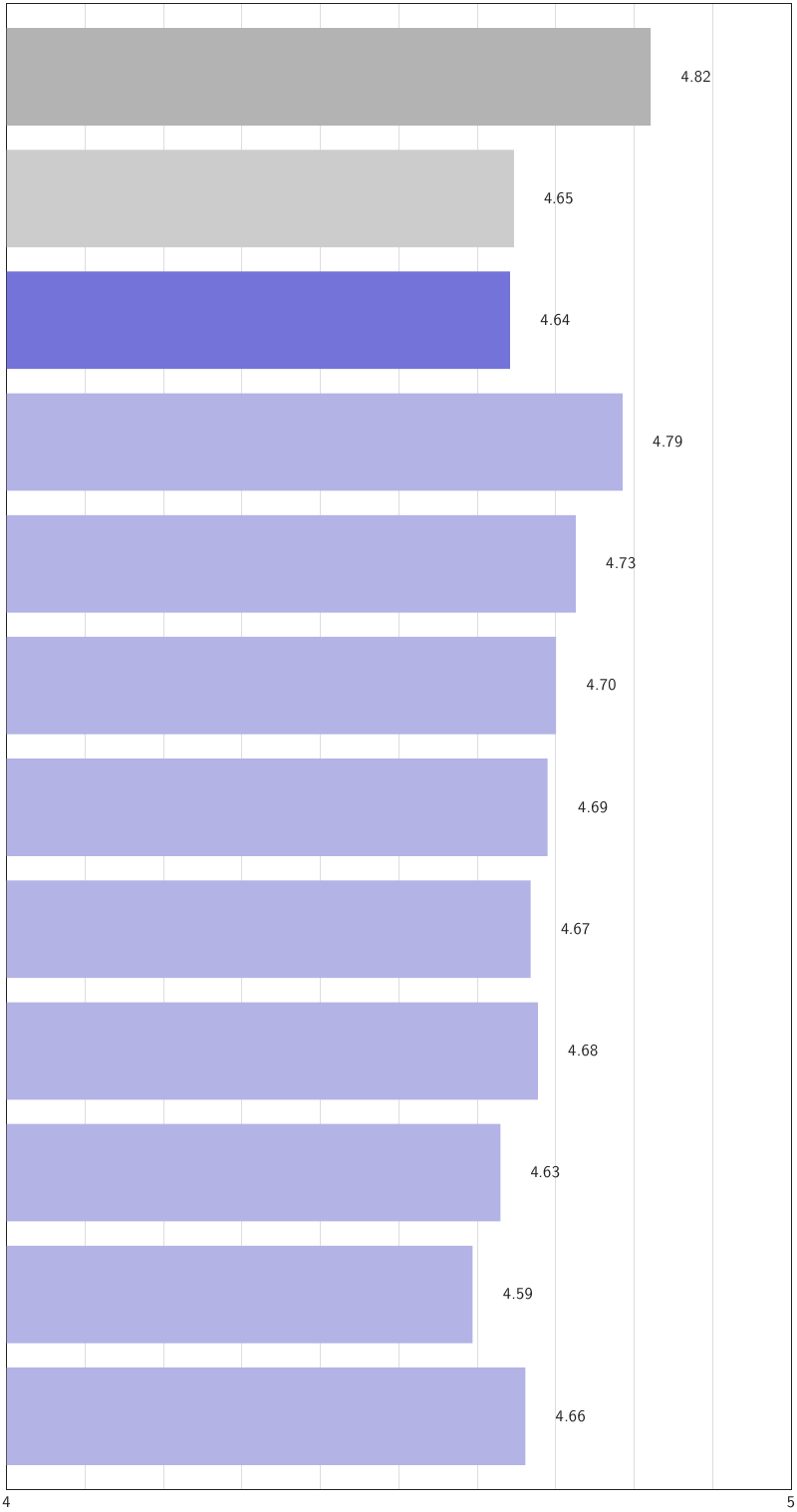}
        %\\
        %\bottomrule
    \end{tabular}
\end{table}%
\vspace{-6pt}

\end{document}